\documentclass[aps, pre, a4paper, floatfix, twocolumn, longbibliography, footinbib, amssymb]{revtex4-2}

\usepackage[colorlinks=true, hypertexnames=false]{hyperref}
\PassOptionsToPackage{linktocpage}{hyperref} 
\hypersetup{
  colorlinks  = true, 
  urlcolor = magenta!80!black,
  linkcolor  = blue!60!black, 
  citecolor  = blue!60 
}

\usepackage[T1]{fontenc}
\usepackage[utf8]{inputenc}
\usepackage{epsfig}
\usepackage[english]{babel}
\usepackage[table]{xcolor}
\usepackage{graphicx}
\usepackage{amsmath}
\usepackage{relsize}
\usepackage[percent]{overpic}
\usepackage{bm}
\usepackage{hyperref}
\usepackage{siunitx}
\usepackage{accents}
\usepackage{mathtools}
\usepackage{grffile}
\usepackage[normalem]{ulem}
\usepackage{pgf}
\usepackage{tikz}
\usepackage{times}

\usepackage[capitalize]{cleveref}

\usepackage{dsfont}  

\newcommand{\avg}[1]{{\left<#1\right>}}
\newcommand{\floor}[1]{{\lfloor #1\rfloor}}

\newcommand{\dd}{\mathrm{d}}
\newcommand{\ee}{\mathrm{e}}

\usepackage{mathtools}
\def\multiset#1#2{\ensuremath{\left(\kern-.3em\left(\genfrac{}{}{0pt}{}{#1}{#2}\right)\kern-.3em\right)}}

\usepackage{amsmath}

\newcommand{\A}{\bm{A}}
\newcommand{\G}{\bm{G}}

\newcommand{\bb}{\bm{b}}

\newcommand{\W}{\bm{W}}
\newcommand{\X}{\bm{X}}

\newcolumntype{P}[1]{>{\centering\arraybackslash}p{#1}}
\usepackage{algpseudocodex}
\usepackage{algorithm}

\algnewcommand\Input{\State\textbf{Input:} }
\algrenewcommand\Output{\State\textbf{Output:} }
\algnewcommand\Continue{\State\textbf{continue}}

\begin{document}

\title{Uncertainty quantification and posterior sampling for network reconstruction}

\author{Tiago \surname{P. Peixoto}}
\email{tiago.peixoto@it-u.at}
\affiliation{Inverse Complexity Lab, IT:U Interdisciplinary Transformation University, 4040 Linz, Austria}

\begin{abstract}
  Network reconstruction is the task of inferring the unseen interactions
  between elements of a system, based only on their behavior or dynamics. This
  inverse problem is in general ill-posed, and admits many solutions for the
  same observation. Nevertheless, the vast majority of statistical methods
  proposed for this task---formulated as the inference of a graphical
  generative model---can only produce a ``point estimate,'' i.e. a single
  network considered the most likely. In general, this can give only a limited
  characterization of the reconstruction, since uncertainties and competing
  answers cannot be conveyed, even if their probabilities are comparable, while
  being structurally different. In this work we present an efficient MCMC
  algorithm for sampling from posterior distributions of reconstructed networks,
  which is able to reveal the full population of answers for a given
  reconstruction problem, weighted according to their plausibilities. Our
  algorithm is general, since it does not rely on specific properties of
  particular generative models, and is specially suited for the inference of
  large and sparse networks, since in this case an iteration can be performed in
  time $O(N\log^2 N)$ for a network of $N$ nodes, instead of $O(N^2)$, as would
  be the case for a more naive approach. We demonstrate the suitability of our
  method in providing uncertainties and consensus of solutions (which provably
  increases the reconstruction accuracy) in a variety of synthetic and empirical
  cases.
\end{abstract}

\maketitle

\section{Introduction}

Many complex systems are governed by interactions that cannot be easily observed
directly. For example, while we can use testing to measure individual infections
during an epidemic spreading, measuring the direct transmission contacts that
caused them are significantly
harder~\cite{netrapalli_learning_2012,braunstein_alfredo_network_2019}. Similarly, we
can measure the abundance of different species in an ecosystem, or the level of
gene expression in a cell, with relatively simple methodologies (e.g. via qPCR
DNA amplification, or DNA microarrays), but determining directly the
interactions between any two species (e.g.\ mutualism or
competition)~\cite{faust_microbial_2012-2,guseva_diversity_2022-1} or any two
genes~\cite{wang_review_2014,pratapa_benchmarking_2020} is significantly more cumbersome. Another
prominent example is the human brain, which can have its behavior harmlessly
probed by an fMRI scan, but its direct neuronal structure cannot be measured
non-invasively. In all these cases, network reconstruction needs to be performed
based on the indirect information available, if we wish to understand how the
system functions.

Several different methods have been proposed for the task of network
reconstruction. A significant fraction of them are heuristic in nature, and
attempt to determine the existence of an edge from pairwise correlations of the
activities of two
nodes~\cite{bullmore_complex_2009,zhang_general_2005,horvath_weighted_2011,tumminello_correlation_2010,zhou_teleconnection_2015,becker_large-scale_2023}.
These methods are fundamentally limited in two important ways. Firstly, they
conflate correlation with conditional dependence or causation, since two nodes
may be strongly correlated even if they are not directly connected (e.g.\ if
they share a neighbor in common). Secondly, with these methods, the existence of
an edge is decoupled from any explicit modelling of the dynamics or behavior of
the system, which severely hinders the interpretability of the reconstruction---
after all, how much would we have really uncovered about a network system if we
do not know how an edge contributes to its
function?~\cite{peel_statistical_2022}. Another prominent class of methods is
based on the definition of explicit generative probabilistic models for the
behavior of a system, conditioned on network of interactions operating as the
parameters of this
model~\cite{dempster_covariance_1972,friedman_sparse_2008,nguyen_inverse_2017,braunstein_alfredo_network_2019}.
In this case, the reconstruction amounts to the statistical inference of these
parameters from data. Within a Bayesian workflow~\cite{gelman_bayesian_2020},
this inferential approach offers a series of advantages, including: 1. A more
principled methodology, coupling tightly theory with data, and relying on
explicit---and hence scrutinizable---modelling assumptions; 2. Non-parametric
implementations~\cite{peixoto_network_2024} dispense with the need to make
\emph{ad hoc} choices, such as arbitrary thresholds, total number of inferred
edges, etc.; 3. The inherent connection with the minimum description length
(MDL) principle~\cite{rissanen_modeling_1978,rissanen_information_2010} provides
a robust framework for model selection~\cite{peixoto_network_2024}, according to
the combined quality of fit and parsimony of the models considered, such that
different hypotheses can be directly compared; and finally, 4. Recent
advances~\cite{peixoto_scalable_2024,peixoto_network_2024} allow for scalable,
sub-quadratic reconstruction of large networks, making the overall approach
practical.

However, despite these advantages, so far the literature on network
reconstruction deals almost exclusively with point estimates, i.e. most of the
methods proposed can only produce a single network, considered to be the most
likely one~\footnote{A notable exception is the literature on reconstruction of
  uncertain or incomplete networks, i.e. when the data is a direct measurement
  of a network, but which has either been corrupted by measurement errors, or
  parts of it have not been measured at all. For this specific class of
  reconstruction problems, posterior sampling and uncertainty quantification is
  more
  commonplace~\cite{butts_network_2003,guimera_missing_2009,newman_network_2018,peixoto_reconstructing_2018,young_bayesian_2021}.
  However, despite both problems sharing the same overall conceptual framework,
  network reconstruction from dynamics or behavior is algorithmically very
  different from the reconstruction of noisy or incomplete networks, and hence
  requires different computational techniques.}, and do not allow for
uncertainty quantification---arguably one of the most desirable and important
features of an inferential analysis. In other words, these point estimates
contain no information about possible alternatives, how different and plausible
they are, and hence how confident we can be about the point estimate in the
first place. Besides this limitation that point estimation imposes on
interpretability, its accuracy is also in general inferior to estimates that
attempt to summarize the consensus over many possible solutions, weighted
according to their plausibility~\cite{jaynes_probability_2003}.

One important reason why point estimation is predominantly employed is its
relative algorithmic efficiency, when compared with approaches based on
posterior averages. This is the main issue we address in this work, where we
develop a scalable algorithm for posterior sampling of reconstructed networks
that performs substantially better for larger problem instances than the naive
baseline. More specifically, whereas a naive implementation of a sampling scheme
would take time $O(N^{2})$ to reconstruct a sparse network of $N$ nodes, our
algorithm is capable of doing the same in time $O(N\log^2 N)$.

This paper is organized as follows. In Sec.~\ref{sec:framework} we describe our
overall inferential framework, and in Sec.~\ref{sec:posterior} our posterior
sampling approach. In Sec.~\ref{sec:synthetic} we compare the performance of
posterior sampling with point estimates for synthetic examples. In
Sec.~\ref{sec:empirical} we do the same for empirical data, where we make also a
comparison with correlation-based reconstructions. We finalize in
Sec.~\ref{sec:conclusion} with a discussion.

\section{Inferential framework}\label{sec:framework}

The inferential scenario for network reconstruction consists of some
data $\X$ that are assumed to originate from a generative model with a
likelihood
\begin{equation}
  P(\X | \W),
\end{equation}
where $\W \in \mathbb{R}^{N\times N}$ is a symmetric matrix corresponding to the
weights of an undirected graph of $N$ nodes (the alternative scenario for
directed networks is straightforward, so we will focus on the undirected case
for simplicity). In most cases we expect $\W$ to be sparse i.e.\ its number of
non-zero entries scales as $O(N)$, but we do not wish to impose any strict
constraints on what values it can take. In many cases, the data are represented
by a $N\times M$ matrix of $M$ i.i.d. samples, with $X_{im}$ being a value
associated with node $i$ for sample $m$, such that
\begin{equation}
  P(\X | \W) = \prod_{m=1}^{M}P(\bm x_{m} | \W),
\end{equation}
with $\bm x_{m}$ being the $m$-th column of $\X$. Alternatively, we may
have that the network generates a Markovian time series with likelihood
\begin{equation}
  P(\X | \W) = \prod_{m=1}^{M}P(\bm x_{m} | \bm x_{m-1},\W),
\end{equation}
given some initial state $\bm x_{0}$. Many other possibilities exist, but for our
purposes we need only to refer to a generic posterior distribution
\begin{equation}\label{eq:posterior}
  P(\W | \X) = \frac{P(\X| \W)P(\W)}{P(\X)},
\end{equation}
which fully quantifies the reconstruction according to some specific generative
model. Since the posterior ascribes a probability to every possible
reconstructed network $\W$, it also quantifies the uncertainty of our inference:
how sharply or broadly ``peaked'' a distribution is around the most likely
network $\W^*$ means that we should have a correspondingly large or small
confidence on its validity as a reconstruction.

Usually, the full posterior distribution is difficult to inspect directly due to
its high-dimensional nature. If we are only interested in a particular
descriptor $f(\W)$ of the reconstructed network, we can avoid this inspection by
computing the posterior mean,
\begin{equation}
  \bar{f}(\X) = \int f(\W)  P(\W | \X)\;\dd\W,
\end{equation}
or, more completely, the marginal posterior distribution
\begin{equation}
  P(y|\X) = \int \delta(y-f(\W))  P(\W | \X)\;\dd\W,
\end{equation}
which fully quantifies the range of plausible descriptor values.

An alternative task is to summarize the posterior distribution as a whole, via a
representative point estimate, and a dispersion around it. There is no unique
way to obtain this summary, which will in general depend on a chosen error
function $\epsilon(\W,\W')$ that we use to evaluate how close is a reconstructed
network $\W$ from the true network $\W'$, with $\W' =
\operatorname{arg\,min}_{\W}\epsilon(\W,\W')$. Since our actual knowledge of the
true network is given by the posterior distribution, we need to consider the
posterior average error
\begin{equation}
  \bar{\epsilon}(\W,\X) = \int \epsilon(\W,\W')P(\W'|\X)\;\dd\W'.
\end{equation}
The representative reconstruction $\widetilde\W$ is the one that minimizes the
average error,
\begin{equation}\label{eq:Wtilde}
  \widetilde\W(\X) = \underset{\W}{\operatorname{arg\,min}}\,\bar{\epsilon}(\W,\X).
\end{equation}
If we choose the maximally strict ``all or nothing'' error function given by
\begin{equation}
  \epsilon(\W,\W') =
  \begin{cases}
    0, &\text{ if } \W = \W',\\
    1, &\text{ otherwise, }
  \end{cases}
\end{equation}
then Eq.~\ref{eq:Wtilde} recovers the maximum \emph{a posteriori} (MAP) point
estimate $\widetilde\W(\X) = \W^{*}(\X)$, with
\begin{equation}\label{eq:MAP}
  \W^{*}(\X) = \underset{\W}{\operatorname{arg\,max}}\,P(\W|\X).
\end{equation}
However, this choice only highlights the lack of nuance the MAP estimator
provides in quantifying uncertainty, since its corresponding error function does
not account for any amount of gradation. Instead, we may wish to account for the
mean squared error
\begin{equation}
  \epsilon(\W,\W') = \sum_{i<j}(W_{ij}-W_{ij}')^{2},
\end{equation}
which provides a gradation not only for the errors of individual entries
$W_{ij}$, but also between all the entries independently. In this case, the
estimator of Eq.~\ref{eq:Wtilde} becomes the pairwise posterior mean,
$\widetilde W_{ij}(\X) = \overline{W}_{ij}(\X)$, with
\begin{equation}
  \overline{W}_{ij}(\X) =  \int W_{ij}'P(\W'|\X)\;\dd\W'.
\end{equation}
Although this estimator may seem entirely reasonable at first, there is still
one remaining issue left to consider. Namely, the scenario we most often expect
to encounter is one where the underlying network $\W$ is sparse, i.e.\ most of
its entries are exactly zero. However, the posterior mean
$\overline{W}_{ij}(\X)$ will not be able to convey sparsity, unless the zeros of
$\W$ occur with absolute certainty in the posterior distribution. Or putting it
differently, the posterior mean alone cannot distinguish between having a high
probability for both zero and non-zero weights, or strictly non-zero weights
distributed with the same mean.

We can address the sparsity estimation by considering an auxiliary
dichotomization $\bm A(\W)$ with entries given by
\begin{equation}
  A_{ij}(\W) =
  \begin{cases}
    1, &\text{ if } W_{ij} \neq 0,\\
    0, &\text{ otherwise, }
  \end{cases}
\end{equation}
and an error function given by
\begin{equation}
  \epsilon(\W,\W') = \sum_{i<j}(W_{ij}-W_{ij}')^{2} + \alpha \left[A_{ij}(\W)-A_{ij}(\W')\right]^{2},
\end{equation}
where $\alpha \geq 0$ denotes the relative importance of the sparsity structure
in the estimation. If we assume $\alpha\to\infty$, the estimator of
Eq.~\ref{eq:Wtilde} becomes $\widetilde W_{ij}(\X) = \widehat{W}_{ij}(\X)$,
with
\begin{equation}\label{eq:mpe}
  \widehat{W}_{ij}(\X) =
  \begin{cases}
    \overline{W}_{ij}(\X),& \text{ if } \pi_{ij}(\X) > \frac{1}{2},\\
    0, &\text{ otherwise, }
  \end{cases}
\end{equation}
where
\begin{equation}\label{eq:pi}
  \pi_{ij}(\X) =  \int A_{ij}(\W')P(\W'|\X)\;\dd\W',
\end{equation}
is the marginal posterior probability of an edge having nonzero weight. We call
the estimator of Eq.~\ref{eq:mpe} simply the ``marginal posterior'' (MP)
estimator from now on. Its uncertainty can be quantified jointly by $\bm\pi(\X)$
and the marginal distributions
\begin{equation}
  P(W_{ij}|\X) =  \int \delta(W_{ij}-W_{ij}')P(\W'|\X)\;\dd\W',
\end{equation}
or more succinctly, by the posterior variances
\begin{equation}
  \sigma_{ij}^{2}(\bm X) = \int \left[W_{ij}'-\overline{W}_{ij}(\X)\right]^{2}P(\W'|\X)\;\dd\W'.
\end{equation}

\subsection{Monte-Carlo sampling}

The above estimators require us to perform posterior means of the type
\begin{equation}
  \int g(\W)  P(\W | \X)\;\dd\W,
\end{equation}
for a particular function $g(\W)$, but exact evaluations of such integrals are
in general intractable. Instead, we need to approximate them as
\begin{equation}
  \int g(\W)  P(\W | \X)\;\dd\W \approx \frac{1}{S}\sum_{k=1}^{S}g(\W^{(k)}).
\end{equation}
where $\{\W^{(1)},\dots,\W^{(S)}\}$ are $S$ samples from the posterior
distribution, which becomes asymptotically exact for $S\to\infty$. The
central surrogate task then becomes to obtain such samples efficiently. We
address the main strategy and its obstacles in the following.

\section{Posterior sampling and the quadratic mixing problem}~\label{sec:posterior}

Our approach for sampling from the posterior of Eq.~\ref{eq:posterior} is to
employ Markov-chain Monte Carlo (MCMC) with the
Metropolis-Hastings~\cite{metropolis_equation_1953,hastings_monte_1970}
acceptance criterion: Given an initial weighted adjacency matrix $\W$, we
propose a new matrix $\W'$ by first selecting a single entry $(i,j)$ of $\W$
with probability $Q(i,j|\W)$, and then changing its value according to a local
proposal $Q(W'_{ij}|i,j,\W)$, and finally accepting the move with probability
\begin{multline}\label{eq:mh}
  a(\W',\W, i,j) = \\\min\left(1, \frac{P(\W'|\X)Q(W_{ij}|i,j,\W')Q(i,j|\W')}{P(\W|\X)Q(W'_{ij}|i,j,\W)Q(i,j|\W)}\right),
\end{multline}
which accounts for the reverse move probability to enforce the detailed balance
condition, given by
\begin{multline}\label{eq:detailed}
  P(\W|\X)Q(W_{ij}'|i,j,\W)Q(i,j|\W)a(\W',\W,i,j) = \\
  P(\W'|\X)Q(W_{ij}|i,j,\W')Q(i,j|\W')a(\W,\W',i,j).
\end{multline}
This condition guarantees that a Markov chain implemented in this way will have
the target posterior $P(\W|\X)$ as its stationary distribution---provided it
exists, i.e. the Markov chain is aperiodic, and as long as the chosen proposal
distributions $Q(i,j|\W)$ and $Q(W'_{ij}|i,j, W_{ij})$ are ergodic, i.e. they allow
for every possible value of the weighted adjacency matrix to be obtained with a
non-zero probability after a finite number of moves.

An appealing property of the MCMC approach is that it obviates the computation
of the usually intractable normalization constant $P(\X)$ that completes the
definition of the posterior distribution $P(\W|\X)$, since this quantity appears
both in the numerator and denominator of Eq.~\ref{eq:mh}, and thus does not
affect the acceptance rate. Therefore, using this scheme, only the joint
likelihood $P(\X,\W)$ is needed to be able to asymptotically sample from the
posterior $P(\W|\X)$.

However, the efficacy of the overall approach hinges crucially on the choice of
the proposal distributions $Q(i,j|\W)$ and $Q(W'_{ij}|i,j, \W)$, since not all
valid choices will lead to the same mixing time, i.e.\ the number of steps needed
to reach the stationary distribution given some initial state. An efficient
proposal distribution will result in fast mixing, allowing for sufficiently many
independent samples from the target distribution to be obtained with relatively
short MCMC runs.

Perhaps the simplest overall scheme is to select the entry $(i,j)$ to be
updated uniformly at random, i.e.
\begin{equation}\label{eq:quniform}
  Q_{u}(i,j) = \frac{\mathds{1}_{i<j}}{{N\choose 2}}.
\end{equation}
Unfortunately, this simple idea will be extremely inefficient in the most
empirically relevant scenarios, even if the local weight proposal $Q(W'|i,j,\W)$
is chosen ideally. This will happen whenever the marginal distribution $\bm\pi$
defined in Eq.~\ref{eq:pi} is sufficiently concentrated on a sparse set of
typical edges, with the remaining entries having $\pi_{ij} < \epsilon$, for some
small probability $\epsilon$. In this case, the total number of typical edges
is given by
\begin{equation}
  |\mathcal{E}| = \sum_{i<j}\mathds{1}_{\pi_{ij} > \epsilon}.
\end{equation}
If, for example, this number grows only linearly as $|\mathcal{E}| = O(N)$, then
the uniform proposal of Eq.~\ref{eq:quniform} will choose an atypical entry
$(i,j)$, i.e.\ one for which $\pi_{ij} < \epsilon$, with a probability
$1-O(1/N)$, hence tending to one for $N\to\infty$. For such atypical entries, a
move that changes its weight from zero to any non-zero value will be accepted
with probability at most $\epsilon$, meaning that the vast majority of moves
will be wasted on vain attempts of placing unlikely edges. In this scenario, the
average time needed to propose a single update to all $|\mathcal{E}|$ typical
edges will scale as $O(N^2)$, which will be a lower bound to the overall mixing
time of the Markov chain.

Instead, an efficient proposal would choose entries according to their
probability to lead to a move being successful. A successful move proposal is
one that combines two properties: 1. It gets accepted; 2. The new value for
$W'_{ij}$ is sufficiently different from the previous value $W_{ij}$---in
particular if $W_{ij} = 0$ then $W'_{ij}\ne 0$, and vice versa. This means that
an efficient entry proposal needs to be able to estimate the typical edge set
---in other words, we need to be able to estimate, beforehand, which entries of
the marginal posterior $\pi$ have sufficiently high values. If this succeeds, we
would be able to update all typical edges in time $O(N)$, significantly reducing
the mixing time when compared to the uniform entry proposal of
Eq.~\ref{eq:quniform}. We describe our approach to achieve this in the following.

\subsection{Estimating the typical edge set}

Our basic idea to estimate the typical edge set is to exploit the information
used to obtain the MAP estimate of Eq.~\ref{eq:MAP}, as described in
Refs.~\cite{peixoto_scalable_2024,peixoto_network_2024}. More specifically, the
algorithm for this purpose consists of iteratively improving the estimate for
$\W^{*}$, starting from an initial $\W = \W^{(0)}$ at $t=0$ containing all
zeros, and proceeding as:
\begin{enumerate}
  \item At iteration $t+1$, given an initial $\W=\W^{(t)}$, we find the set
        $\mathcal{E}^{(t+1)}$ containing the $\kappa N$ entries of $\W$ that
        most increase or least decrease the posterior $P(\W|\X)$, with $\kappa$
        being a parameter of the algorithm.
  \item The entries of $\mathcal{E}^{(t+1)}$ are updated in sequence to maximize $P(\W|\X)$, yielding a new estimate $\W^{(t+1)}$.
  \item If the difference between $\W^{(t+1)}$ and $\W^t$ falls below some
    tolerance value, we return $\W^* = \W^{(t+1)}$, otherwise we continue
    from step 1.
\end{enumerate}
A naive implementation of step 1 would exhaustively search through all entries,
taking time $O(N^2)$. Instead, as described in
Ref.~\cite{peixoto_scalable_2024}, it is possible to estimate
$\mathcal{E}^{(t+1)}$ in subquadratic time, typically $O(\kappa^{2}N\log^{2}N)$,
using a recursive second-neighbor search. Our estimate $\widehat{\mathcal{E}}$
for the typical set is then the union of all candidate entries encountered
during the above algorithm, i.e.
\begin{equation}
  \widehat{\mathcal{E}} = \bigcup\limits_{t=1}^{T}\mathcal{E}^{(t)},
\end{equation}
where $T$ is the total number of iterations. Note that we are not interested
only in the last set of candidate edges, nor in the nonzero entries of the final
MAP estimate $\W^{*}$, since we want edges with a non-negligible marginal
probability, not only the most likely ones.

Since $T$ is typically a constant with respect to $N$, the total size of the
typical set is $\widehat{\mathcal{E}}=O(N)$. With our estimate $\widehat{\mathcal{E}}$ at
hand, we propose entries for the MCMC according to
\begin{equation}\label{eq:proposal_mixed}
  Q(i,j) = \frac{w_t Q_{t}(i,j|\hat{\mathcal{E}}) + w_u Q_{u}(i,j)}{w_t + w_u},
\end{equation}
with
\begin{equation}\label{eq:typical}
  Q_{t}(i,j|\hat{\mathcal{E}}) = \frac{\mathds{1}_{(i,j)\in\hat{\mathcal{E}}}}{|\hat{\mathcal{E}}|}.
\end{equation}
and with $w_t$ and $w_u$ being the relative propensities of choosing entries in
the set $\widehat{\mathcal{E}}$ and uniformly, respectively. Note that we need
$w_u>0$ to guarantee ergodicity, but we expect $Q_{t}(i,j|\hat{\mathcal{E}})$ to
yield the most successful proposals.

The above algorithm does not guarantee that all members of the typical set are
found. To increase our chances of finding the entire set, we initialize the MCMC
with the MAP estimate $\W^*$, and after a sweep comprised of $N$ consecutive
proposals, we compute a new set $\mathcal{E}'$ according to the same algorithm
used in step 1 of the above algorithm, and add it to our typical set estimate
$\widehat{\mathcal{E}}$. Note that since this changes the proposal probabilities
that depend on $\widehat{\mathcal{E}}$, this procedure will invalidate detailed
balance, and therefore will not lead to a correct sampling of the target
distribution. Because of this, we perform this update only for $\tau$ initial
sweeps, and afterwards we continue sampling with final set
$\widehat{\mathcal{E}}$ fixed.

\begin{figure}
  \begin{overpic}[width=\columnwidth]{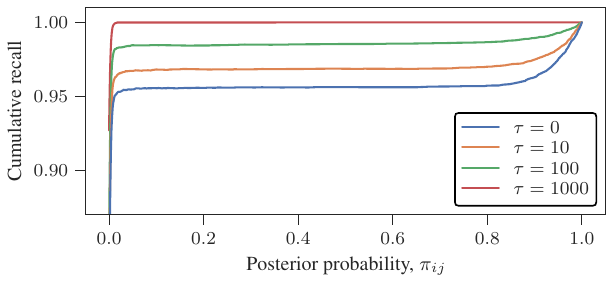}\put(0,45){(a)}\end{overpic}
  \begin{overpic}[width=\columnwidth]{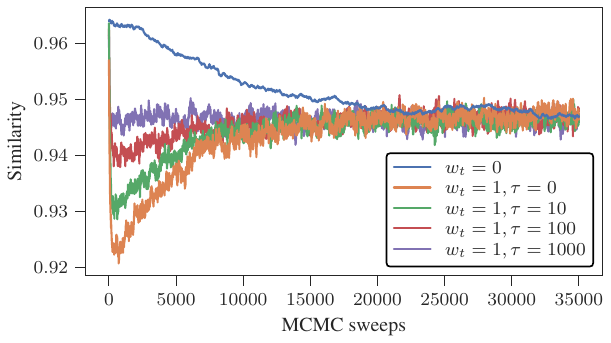}\put(0,55){(b)}\end{overpic}
  \begin{overpic}[width=\columnwidth]{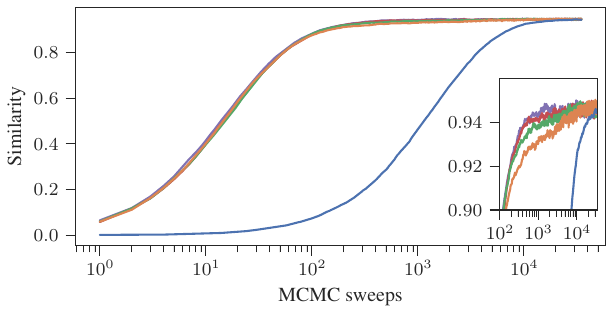}\put(0,50){(c)}\end{overpic}
  \begin{overpic}[width=\columnwidth]{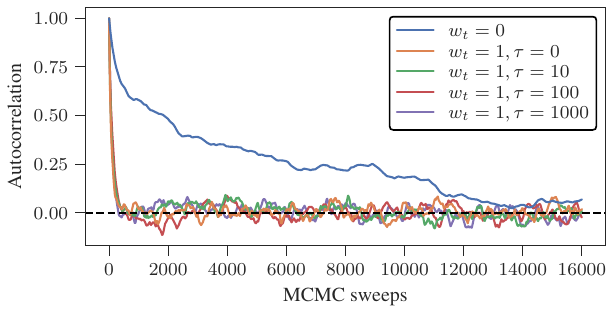}\put(0,50){(d)}\end{overpic}
  \caption{Results of MCMC runs for the reconstruction of an Ed\H{o}s-Rényi
    network of $N=5000$ nodes and average degree $2E/N = 5$, and weights sampled
    from a normal distribution with mean $1/5$ and standard deviation $0.01$,
    serving as the couplings of a kinetic Ising model (see
    Appendix~\ref{app:models}), based on $M=500$ parallel transitions from a
    random initial state. Panel (a) shows the cumulative recall of the typical
    set, i.e.\ the fraction of all entries with a posterior probability
    $\pi_{ij}$ above a particular value that have been found in
    $\hat{\mathcal{E}}$, for several values of the search period $\tau$. Panel
    (b) shows the Jaccard similarity $s(\W',\W)$ between samples $\W'$ generated
    by the MCMC and the true value $\W$, with ($w_t=1$) and without ($w_t=0$)
    the estimation of the typical edge set, and various search periods $\tau$.
    Panel (c) shows the same kinds of MCMC runs, but with an initial state
    consisting of an empty network (the inset shows a zoom in the high
    similarity region). Panel (d) shows the autocorrelation function for the
    values of similarity of the runs in panel (b), discarding the initial
    transient before equilibration.\label{fig:typical-set}}
\end{figure}

In Fig.~\ref{fig:typical-set} we demonstrate the behavior of this algorithm on a
the reconstruction of an Ed\H{o}s-Rényi network of $N=5000$ nodes and average
degree $2E/N = 5$, and weights sampled from a normal distribution with mean
$1/5$ and standard deviation $0.01$, serving as the couplings of a kinetic Ising
model (see Appendix~\ref{app:models}), after $M=500$ parallel transitions from a
random initial state. Fig.~\ref{fig:typical-set}a shows the cumulative recall of
the typical set, i.e.\ the fraction of all entries with a posterior probability
$\pi_{ij}$ above a particular value that have been found in $\hat{\mathcal{E}}$,
for several values of the search period $\tau$. Although for $\tau=0$ the recall
is already $95\%$ for the entire range of typical posterior probabilities, it
increases continuously to $100\%$ for $\tau=10^3$, indicating that further
posterior samples can improve the estimate of the initial greedy algorithm. In
Fig.~\ref{fig:typical-set}b is shown the evolution of the Jaccard similarity
\begin{equation}
  s(\W',\W) = 1 - \frac{\sum_{i<j}|W_{ij}'-W_{ij}|}{\sum_{i<j}|W_{ij}'+W_{ij}|},
\end{equation}
between samples $\W'$ generated by the MCMC and the true value $\W$. Despite the
search time $\tau$ being barely visible in the time-span considered, its
longer-term effect is noticeable, since the MCMC run with $\tau = 10^3$
converges significantly faster than the one with $\tau=0$, despite the
cumulative recall being already $95\%$ in the latter case. This is due to the
fact that the remaining $5\%$ of the typical edge set needs to be found by
uniform sampling, which will still takes an $O(N)$ number of sweeps. In the same
figure we also show the result with $w_t=0$, i.e. using only uniform entry
samples, which displays a much slower convergence. In
Fig.~\ref{fig:typical-set}c we show the results of the same algorithms, but
starting from an empty network (i.e. all entries being zero), where we can see
that the uniform sampling takes a time at least two order of magnitude larger to
converge. Finally, in Fig.~\ref{fig:typical-set}c we show the autocorrelation
function of the similarity, discarding the transient towards equilibration, for
the same runs as before. The runs with $w_t=1$ yield autocorrelation times
ranging from $300$ ($\tau=10^3$) to $600$ ($\tau=0$) sweeps, whereas runs with
$w_t=0$ have an significantly higher autocorrelation time of around $21,000$
sweeps. This demonstrates how this scheme can have a substantial impact on the
efficiency of drawing samples from the posterior distribution via MCMC.

\begin{figure}
  \includegraphics[width=.7\columnwidth]{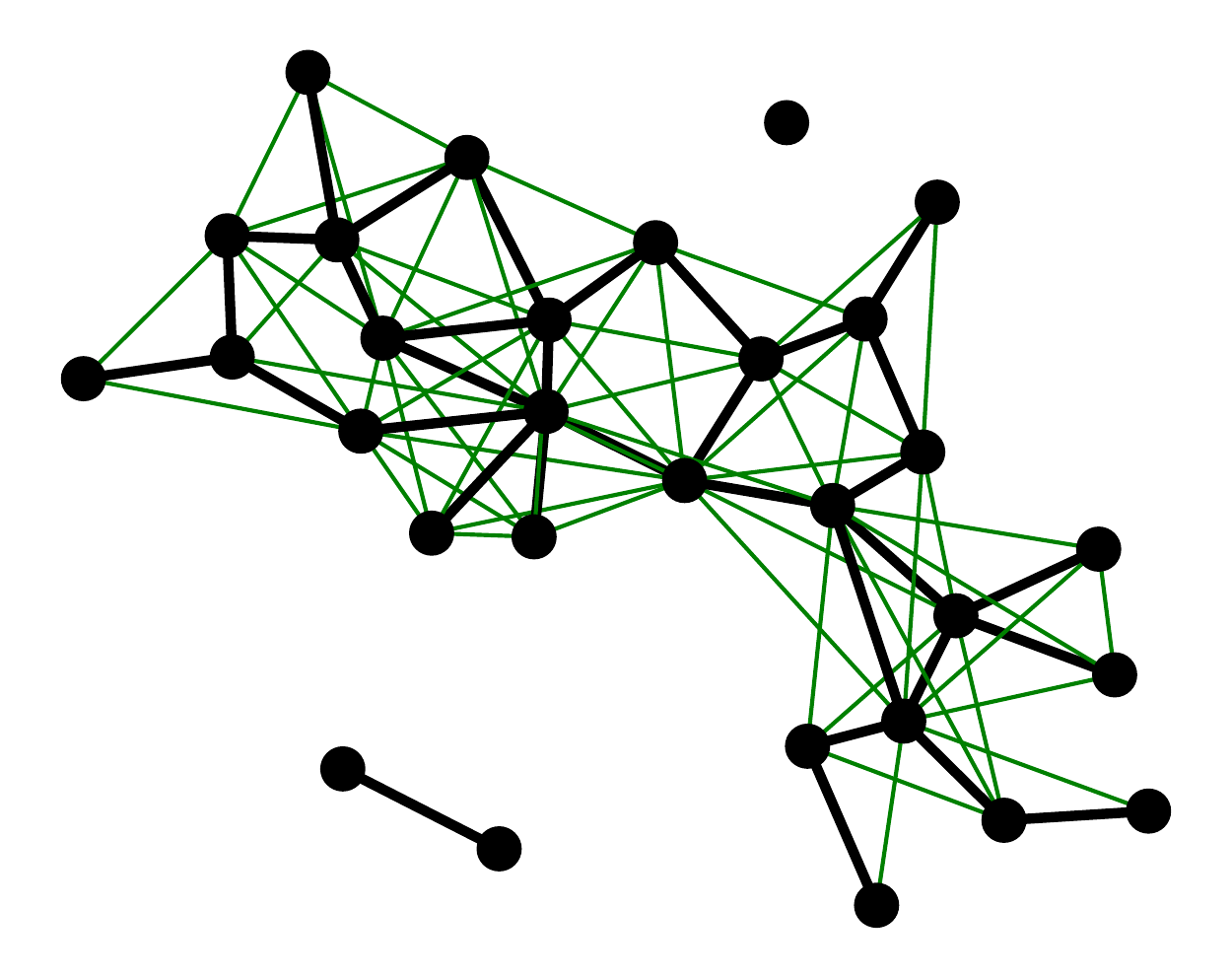}
  \caption{Illustration of the proposed ``nearby'' updates according to
    Eq.~\ref{eq:nearby}. The black edges correspond to the nonzero entries of
    $\W$ at some point of the algorithm, and the green edges are entries
    with $Q_{n}(i,j|\W,d)>0$ for $d=2$, which would be
    proposed for an update. Edges between the different components will never be
    proposed for any value of $d$.\label{fig:nearby}}
\end{figure}

\begin{figure}
  \begin{tabular}{cc}
    Connected typical network & Disconnected typical network \\
    \begin{overpic}[width=.5\columnwidth]{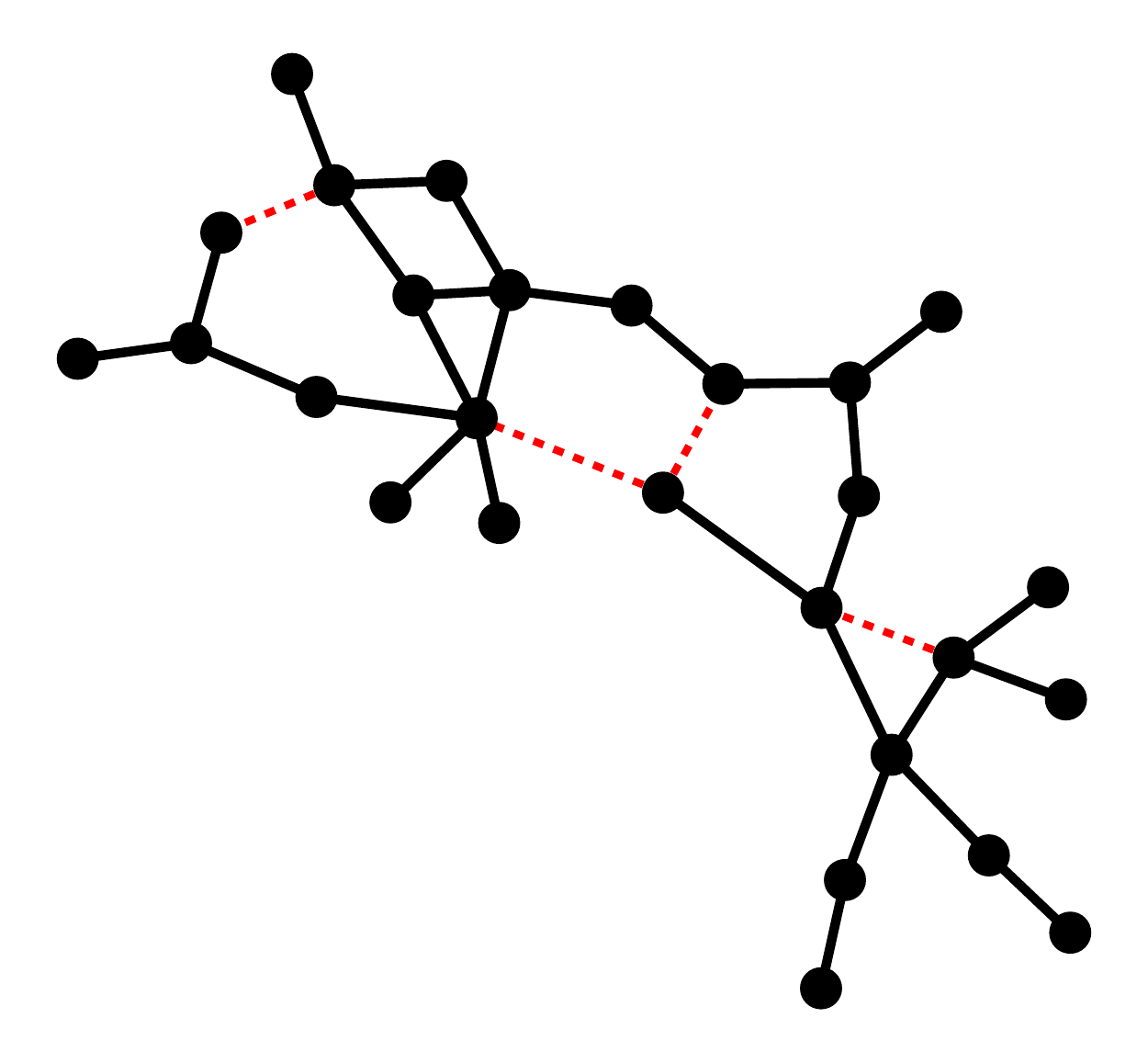}\put(0, 0){(a)}\end{overpic} & \begin{overpic}[width=.5\columnwidth]{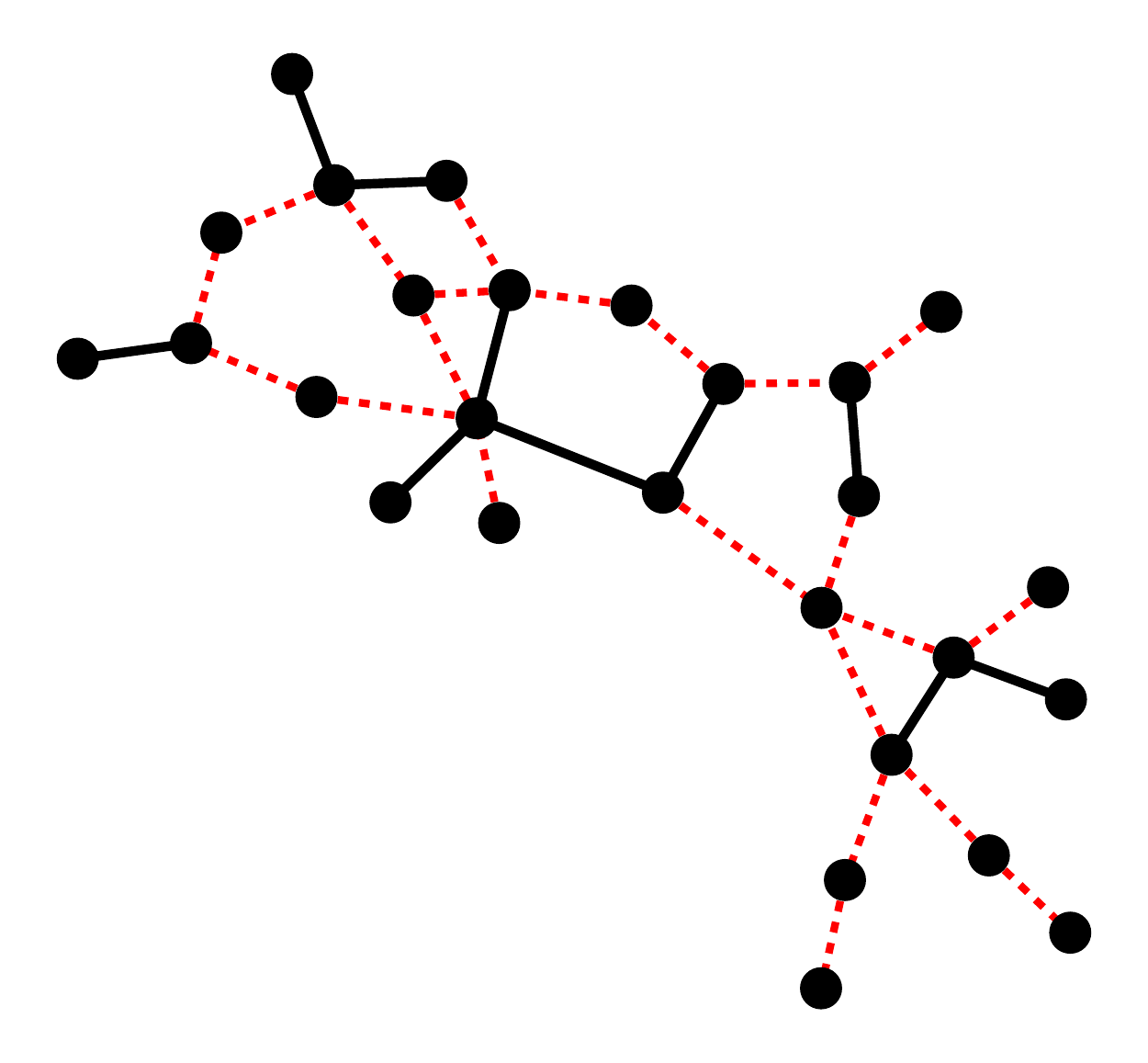}\end{overpic} \\
    \multicolumn{2}{c}{\begin{overpic}[width=\columnwidth]{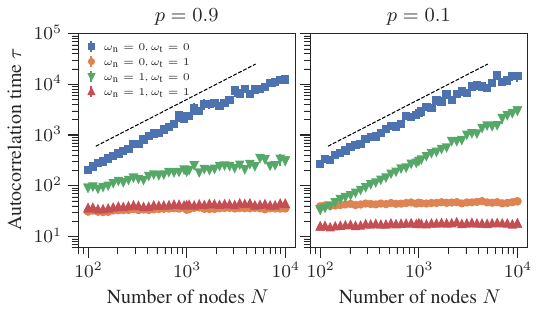}\put(0,0){(b)}\end{overpic}}
  \end{tabular}
  \caption{Panel (b) shows the autocorrelation time as a function of the number
    of nodes $N$, for a target distribution according to Eq.~\ref{eq:target},
    with $\bm G$ generated as described in the text, with $E=5 N / 2$ edges, and
    considering different combinations of the move proposals, as indicated in
    the legend, in the situation where the typical network is connected
    ($p=0.9$) and where it is disconnected ($p=0.1$), in both cases with
    $\epsilon=10^{-8}$. The dashed line indicates a linear slope. Panel (a)
    shows an illustration of the connected and disconnected cases, with black
    edges representing those in $\bm G$ that are currently being sampled, and
    the dashed edges those in $\bm G$ that are not.\label{fig:scaling}}
\end{figure}

\subsubsection{Searching for ``nearby'' edges}

The protocol described previously relies on a pre-processing phase aimed at
determining the typical edge set, before the MCMC proper is run. Here we present
and evaluate an additional strategy which aims to continuously improve our
estimate on the typical edge set during the MCMC, which consists of selecting
preferentially entries that are ``close'' to the current edges of the network
(i.e.\ the nonzero entries of the current state of the MCMC). More specifically,
we choose a node $i$ uniformly at random, and the second node $j$ uniformly from
the set that is reachable from $i$ in the dichotomized network $\bm A(\bm W)$ at
a distance at most $d$, i.e.
\begin{equation}
  R(i,j|\W,d) =
  \begin{cases}
    \frac{\mathds{1}_{j \in \Lambda(i, d)}}{|\Lambda(i, d)|N}, \text{ if } |\Lambda(i, d)| > 0,\\
    \frac{1}{N(N-1)}, \text{ otherwise,}
  \end{cases}
\end{equation}
where $\Lambda(i, d)$ is the set of nodes in $\A(\W)$ that are reachable from
$i$ at a distance at most $d$. Note that in general this proposal is asymmetric,
$R(i,j|\W,d) \neq R(j,i|\W,d)$, so the final probability becomes
\begin{multline}\label{eq:nearby}
  Q_{n}(i,j|\W,d) =\mathds{1}_{i<j}\left[R(i,j|\W,d) + R(j,i|\W,d)\right].
\end{multline}
By itself this proposal will not lead to an ergodic Markov chain, so it needs to
be used together with the proposal of Eq.~\ref{eq:proposal_mixed}.

\begin{figure}
  \begin{tabular}{cc}
    \hspace{3.0em}American college football & \hspace{1em}High school friendships \\
    \multicolumn{2}{c}{\includegraphics[width=\columnwidth]{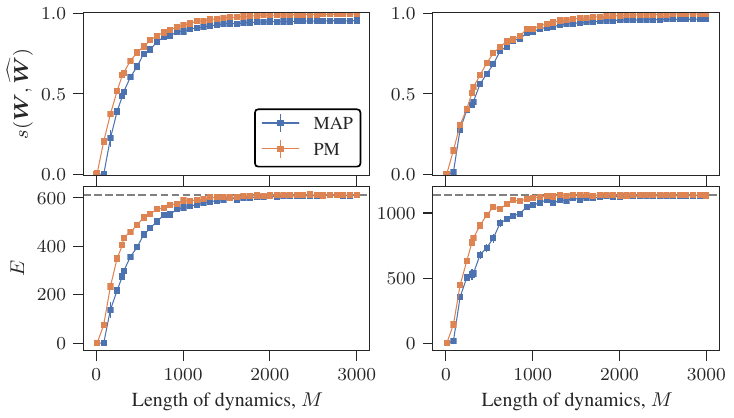}}
  \end{tabular}
  \caption{Reconstruction performance based on the dynamics generated by the
    kinetic Ising model (see Appendix~\ref{app:models}) on two empirical
    networks, where the weights are sampled from a normal distribution with mean
    $1/\avg{k}$ and standard deviation $0.01$, with $\avg{k}=2E/N$ being the
    average degree. The left panels show the results for a network of American
    football teams~\cite{girvan_community_2002} (with $N=115$ and $E=613$), and
    on the left for a network of friendship between high school
    students~\cite{moody_peer_2001} (with $N=291$ and $E=1136$). The panels on
    the top show the similarity $s(\W,\hat{\W})$ between the inferred and true
    networks, according to the MAP and MP estimators, as indicated in the
    legend, as a function of the lenght $M$ of the dynamics. The bottom panels
    show the number of edges of the inferred networks in each case. The dashed
    horizontal lines indicate the true value.\label{fig:synthetic}}
\end{figure}

\begin{figure*}
  \begin{tabular}{ccc}    \multicolumn{3}{c}{\begin{overpic}[width=\textwidth]{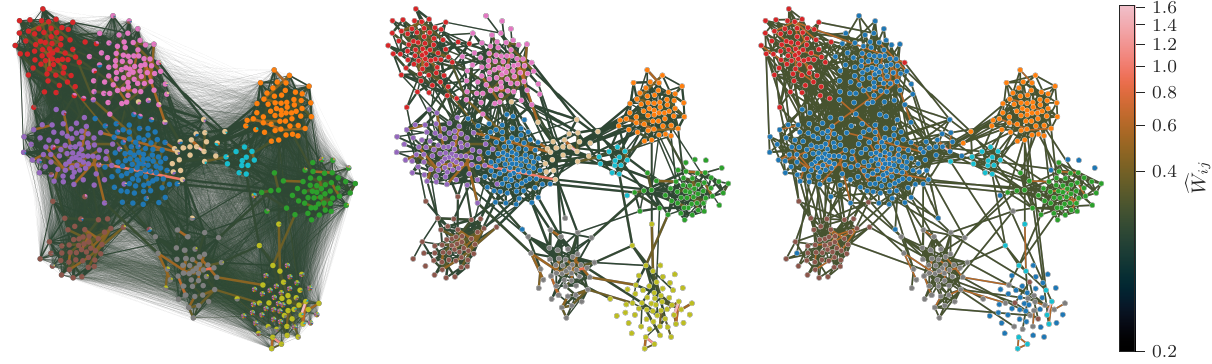}\put(0,30){(a)}\put(30,30){(b)}\put(60,30){(c)}\end{overpic}}\\
    \begin{overpic}[width=.32\textwidth]{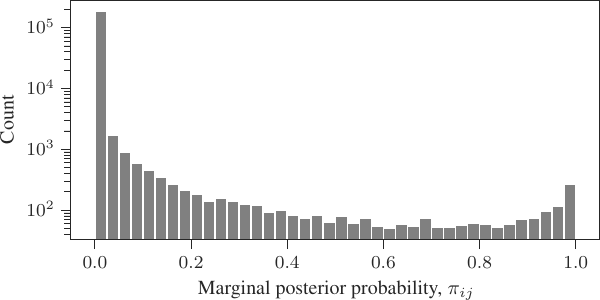}\put(0,50){(d)}\end{overpic} &
    \begin{overpic}[width=.32\textwidth]{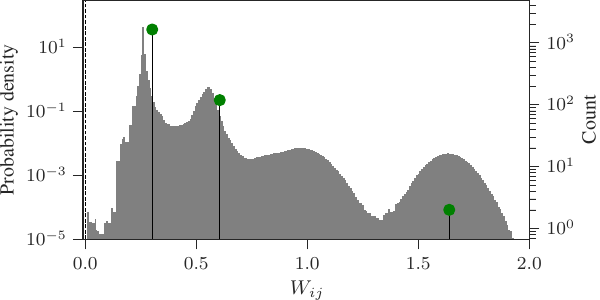}\put(0,50){(e)}\end{overpic} &
    \begin{overpic}[width=.32\textwidth]{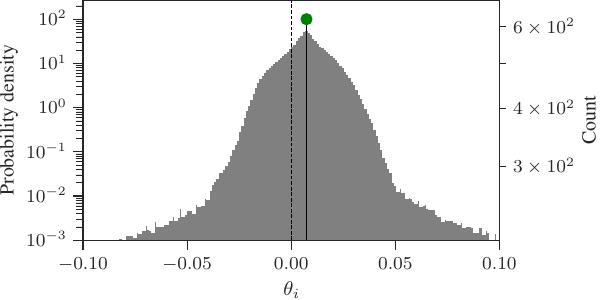}\put(0,50){(f)}\end{overpic}
  \end{tabular}
  \caption{Reconstruction of a zero-added Ising model based on $M=619$ votes of
    $N=623$ deputies of the lower house of the Brazilian congress. (a) Marginal
    edge probabilities $\bm\pi$ indicated as edge thickness and the posterior
    mean $\widehat\W$ as edge colors. The node pie charts indicate the marginal
    group memberships, inferred according to the SBM incorporated in the
    reconstruction, as described in Ref.~\cite{peixoto_network_2024}. (b) MP
    estimate $\widehat\W$ according to Eq.~\ref{eq:mpe}. (c) MAP point estimate
    $\W^{*}$ according to Eq.~\ref{eq:MAP}. (d) Distribution of marginal
    posterior probability values $\pi_{ij}$ across all node pairs. (e) Posterior
    distribution of non-zero weight values $W_{ij}$ across all node pairs. (f)
    Distribution of node biases $\theta_{i}$ across all nodes $i$. In (e) and
    (f) the vertical lines correspond to the distribution obtained with the MAP
    point estimate.\label{fig:camara}}
\end{figure*}

An illustration of the entries that are preferentially sampled in this manner is
shown in Fig.~\ref{fig:nearby}. The intuition behind this idea is that if the
edges of $\A(\W)$ are already in the typical edge set $\mathcal{E}$, then the
entries connecting indirect neighbors are likely to be in this set as well. This
should happen with reconstruction problems with some degree of transitivity,
i.e.\ when direct connections and those between second and third neighbors of
their endpoints might have comparable or at least decaying posterior probabilities.

This approach will fail in two scenarios: 1. When the transitivity property is
not applicable; 2. When the current graph $\A(\W)$ is sufficiently disconnected,
such that entries between different components are never preferentially
proposed. We illustrate the behavior of this kind of proposal on a target
distribution given by $\hat{\bm\pi} = \prod_{i<j}\hat{\pi}_{ij}$, with
\begin{equation}\label{eq:target}
  \hat\pi_{ij} = p^{G_{ij}}\epsilon^{1-G_{ij}},
\end{equation}
where $\bm G$ is a random graph with an increased abundance of triangles,
generated by first sampling an Erd\H{o}s-Rényi network with $E$ edges, removing
$E n / (n+1)$ edges uniformly at random, and then employing the following
procedure $n$ times in succession: Of all open triads in $\bm G$---i.e.\ entries
$(i,j)$ such that $G_{ij}=0$ and $G_{iu}G_{uj}=1$ for some node
$u \notin \left\{i,j\right\}$---$E / (n+1)$ of them are selected uniformly at
random and closed, i.e. $G_{ij}\to 1$. This guarantees that the final graph will
have exactly $E$ edges, and a significantly higher fraction of triangles than
would be expected in an ER network. In Fig.~\ref{fig:scaling} we show the
autocorrelation time for our proposed MCMC as a function of the number of nodes
$N$, for $E=5 N / 2$, considering different combinations of the move proposals
so far considered, in the situation where the typical network is connected
($p=.9$) and where it is disconnected ($p=.1$). In the connected case, the
nearby moves have no noticeable effect on the autocorrelation time when the
initial estimate of the typical edge set is being used $(w_t=1)$, but it
improves significantly the mixing when it is used on its own (in addition to the
uniform moves)---in this case the autocorrelation does not grow linearly with
$N$ as in the case of using only uniform proposals. In the disconnected case, as
expected, the nearby moves lose significantly their efficacy: when used on their
own, the autocorrelation also increases linearly with $N$. However, even in this
case, its use reduces the mixing time by a constant factor, even when combined
with the initial estimation of the typical set. This approach is, therefore,
potentially useful in situations where the typical edge set cannot be accurately
estimated with the protocol described previously.

\subsubsection{Edge weights, node values, and community structure}

In the previous sections we have focused on the move proposals $P(i,j|\W)$ that
involve the selection of entries in the matrix $\bm W$ to be updated, but not on
the proposals $P(\W'|i,j,\W)$ to update the actual value of the entry selected,
since the former is the most crucial for the algorithmic performance. For the
value updates, conventional choices can in principle be used, such as sampling
from a normal distribution. In Appendix~\ref{app:sample_x} we describe an
alternative approach based on bisection sampling that we found to be efficient,
and also works well with regularization schemes that rely in discretization,
such as the minimum description length (MDL) formulation of
Ref.~\cite{peixoto_network_2024}, which we summarize in Appendix~\ref{app:mdl}.

One feature of the MDL regularization is that it includes the stochastic block
model~\cite{holland_stochastic_1983} as a prior, and therefore it performs
community detection as part of the reconstruction, which has been shown
previously to improve the overall accuracy~\cite{peixoto_network_2019}.

Furthermore, most models also include an additional set of parameters $\bm\theta$
on the nodes, that also need to be updated. We have not included these parameters
in our discussion so far, since they can be handled completely separately, by
selecting one of them at random, and using the same kinds of updates as used for
the entries of $\bm W$. Differently from $\bm W$, there is no inherent
algorithmic challenge in sampling these node parameters, since their number
scales only linearly with the number of nodes.

Finally, in Appendix~\ref{app:swap} we also describe an extension of the
algorithm which allows for edge replacements and swaps, that can potentially
move across likelihood barriers present when discretized regularization schemes
are used.

We provide a reference C++ implementation of the algorithms described here,
together with documentation, as part of the \texttt{graph-tool} Python
library~\cite{peixoto_graph-tool_2014}.

\begin{figure}
  \begin{tabular}{c}
    \begin{overpic}[width=\columnwidth]{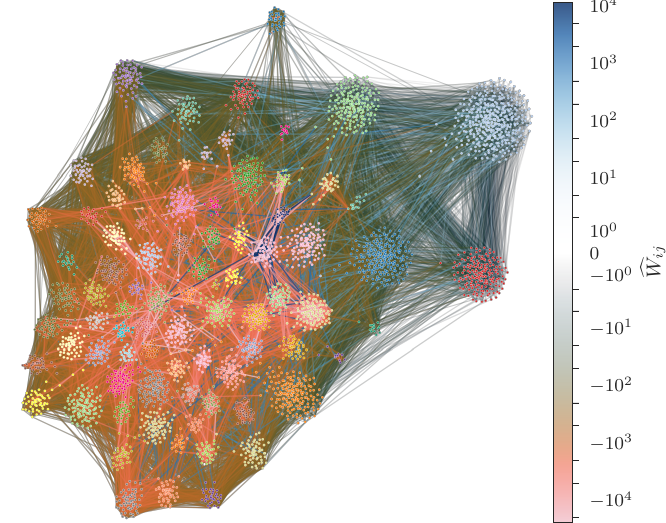}\put(0,73){(a)}\end{overpic}\\[.5em]
    \begin{overpic}[width=\columnwidth]{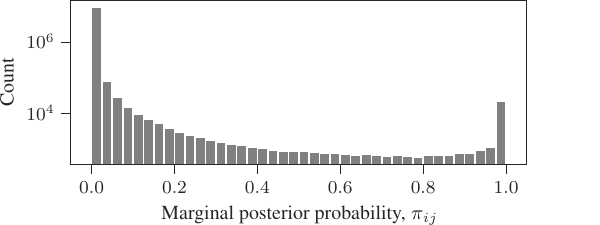}\put(0,38){(b)}\end{overpic}\\
    \begin{overpic}[width=\columnwidth]{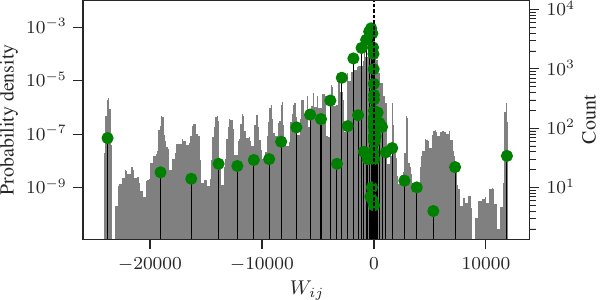}\put(0,50){(c)}\end{overpic}
  \end{tabular}
  \caption{\label{fig:stock}Reconstruction of a multivariate Gaussian model based on $M=2516$ log-returns of
    $N=6369$ US stocks in the period between 2014 to 2024. (a) Marginal
    edge probabilities $\bm\pi$ indicated as edge thickness and the posterior
    mean $\widehat\W$ as edge colors. The node colors indicate the maximum marginal
    group memberships, inferred according to the SBM incorporated in the
    reconstruction, as described in Ref.~\cite{peixoto_network_2024}. (b) Distribution of marginal posterior
    probability values $\pi_{ij}$ across all node pairs. (c) Posterior
    distribution of non-zero weight values $W_{ij}$ across all node pairs. The
    vertical lines correspond to the distribution obtained with the MAP point
    estimate.}
\end{figure}

\begin{figure}
  \begin{tabular}{c}
    (a) Brazilian congress \\[2em]
    \begin{overpic}[width=\columnwidth]{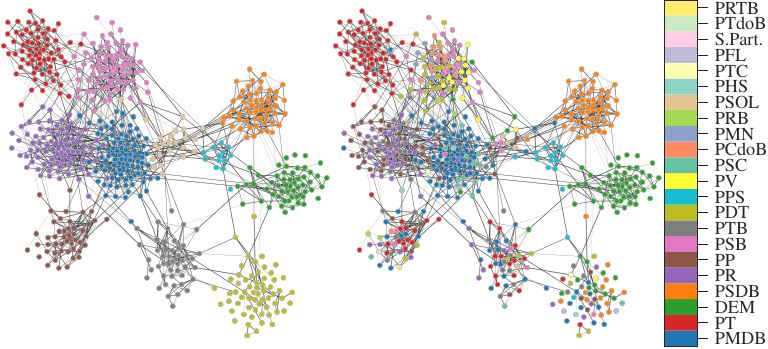}
      \put(8,47){Inferred partition}
      \put(52,47){Party affiliation}
    \end{overpic}\\
    (b) US stock prices \\[2em]
    \begin{overpic}[width=\columnwidth]{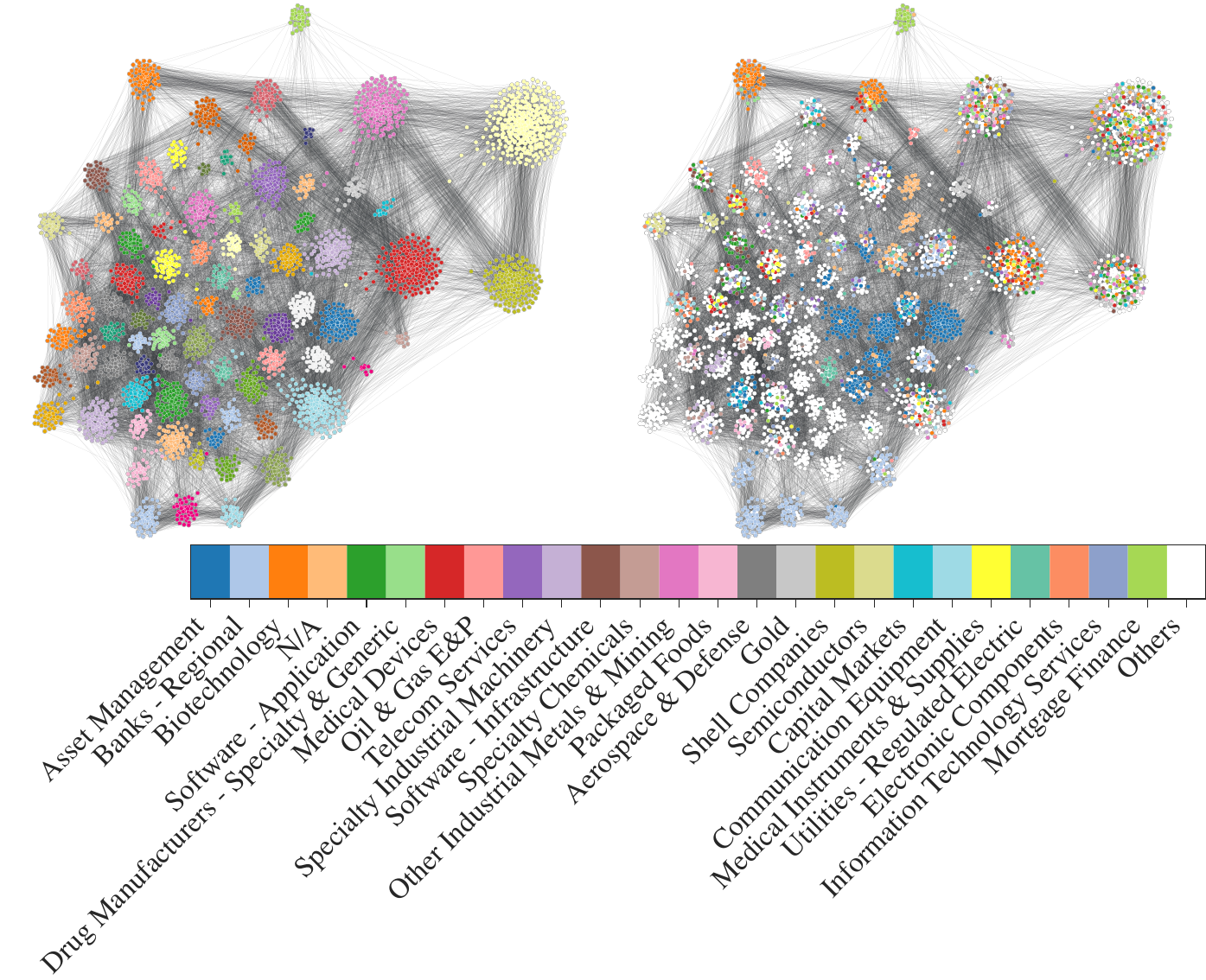}
      \put(10,84){Infered partition}
      \put(65,84){Industry sector}
    \end{overpic}
  \end{tabular}
  \caption{\label{fig:meta} Correspondence between the inferred partition using
    the bult-in SBM in our reconstruction (left) with available metadata on the
    nodes (right), for (a) the Brazilian congress, with the metadata being the
    party affiliation of the deputies, and (b) US stock prices, with the
    metadata being the industrial sector, in both cases as indicated in the
    legend.}
\end{figure}

\begin{figure}
  \begin{tabular}{c}\smaller
    (a) Brazilian congress\\[.5em]
    \includegraphics[width=\columnwidth]{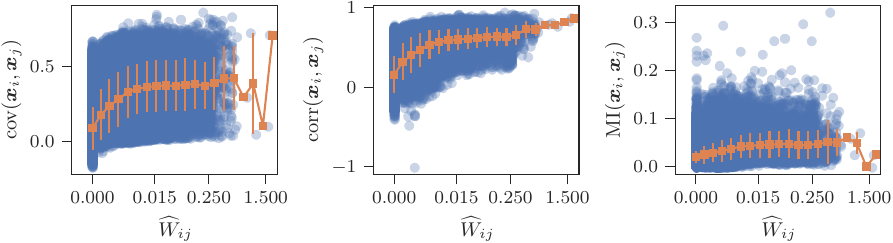}\\
    \includegraphics[width=\columnwidth]{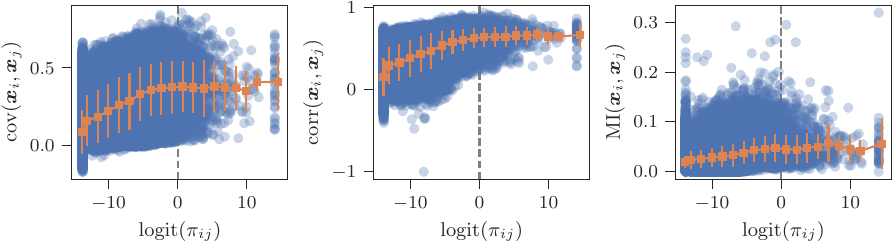}\\\smaller
    (b) US stock prices\\[.5em]
    \includegraphics[width=\columnwidth]{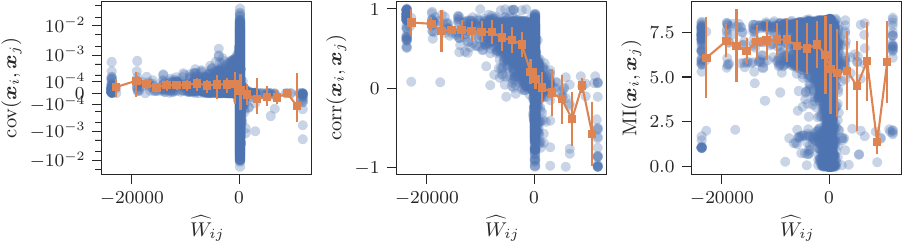}\\
    \includegraphics[width=\columnwidth]{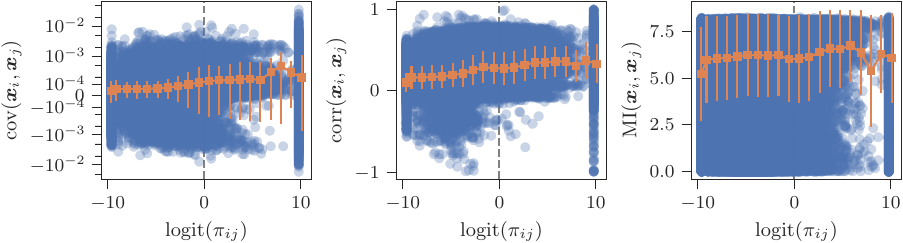}
  \end{tabular}
  \caption{Scatter plot between mean posterior weights $\widehat{W}_{ij}$ or
    posterior probabilities $\pi_{ij}$ and a type of pairwise correlation, i.e.
    either the covariance $\operatorname{cov}(\bm x_{i}, \bm x_{j})$, Pearson
    correlation $\operatorname{corr}(\bm x_{i}, \bm x_{j})$, or mutual
    information $MI(\bm x_{i}, \bm x_{j})$, for every node pair $(i,j)$, for (a)
    the Brazilian congress data, and (b) the US stock prices data. The connected
    orange points correspond to binned averages.\label{fig:scatter}}
\end{figure}

\begin{figure}
  \begin{tabular}{cc}\smaller
    \hspace{3em}(a) Brazilian congress & \smaller \hspace{2.5em} (b) US stocks \\
    \includegraphics[width=.5\columnwidth]{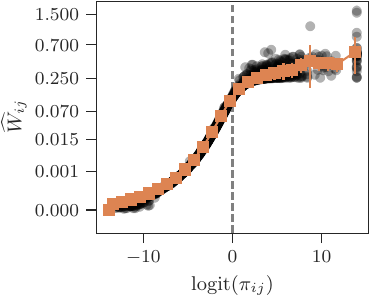} &                                                                                                                \includegraphics[width=.5\columnwidth]{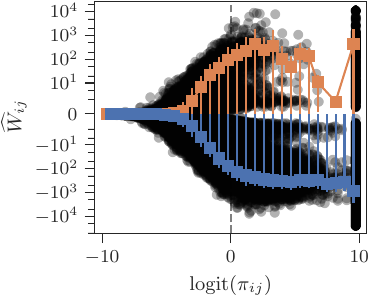}
  \end{tabular}
  \caption{Scatter plot of mean posterior weights $\widehat W_{ij}$ vs.\ posterior
    probabilities $\pi_{ij}$, for every node pair $(i,j)$, for (a) the Brazilian
    congress data, and (b) the US stock prices data. The connected orange points
    correspond to binned averages for positive weights, and the blue points for
    negative weights.\label{fig:scatter_wp}}
\end{figure}

\section{MAP vs MP estimation with synthetic dynamics}\label{sec:synthetic}

In Fig.~\ref{fig:synthetic} we show a comparison between the MAP and MP
estimates for synthetic dynamics, i.e. $M$ transitions of the kinetic Ising
model, on empirical networks, using the MDL regularization of
Ref.~\cite{peixoto_network_2024}, described in Appendix~\ref{app:mdl}. For
sufficient data, both estimates yield the same reconstruction. However, as data
become more scarce, the MP estimator shows a systematically better performance,
although the difference is not very large in these examples. The difference in
performance is unsurprising, given that the derivation of the MP estimator
results from the optimization of the mean squared error, as we discussed
previously, and therefore it cannot be surpassed by MAP.\@ Nevertheless, it
serves as a good demonstration that obtaining the consensus over the posterior
distribution can improve the accuracy of point estimates.

Besides the increased accuracy, posterior estimation can provide uncertainty
quantification. We focus on this aspect when analysing the reconstruction based
on empirical dynamics, in the following.

\begin{figure}
  \begin{tabular}{cc}
    \includegraphics[width=.5\columnwidth]{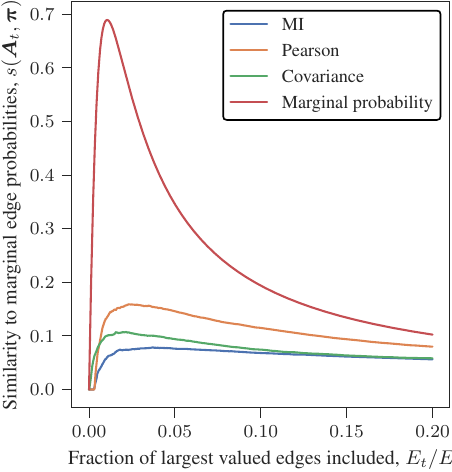} &
    \includegraphics[width=.5\columnwidth]{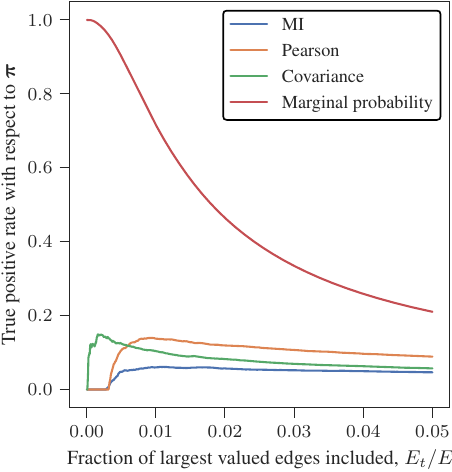}
  \end{tabular}
  \caption{Accuracy according to the fraction of largest values included in the
    reconstruction, for the Brazilian congress data, for different kinds of
    ``scores'' attributed to the edge pairs. The left plot shows the Jaccard
    similarity, while the right shows the ``true positive'' rate, taking the
    marginal probability as reference. \label{fig:corr_acc}}
\end{figure}

\begin{figure}
  \begin{tabular}{c}
    \includegraphics[width=\columnwidth]{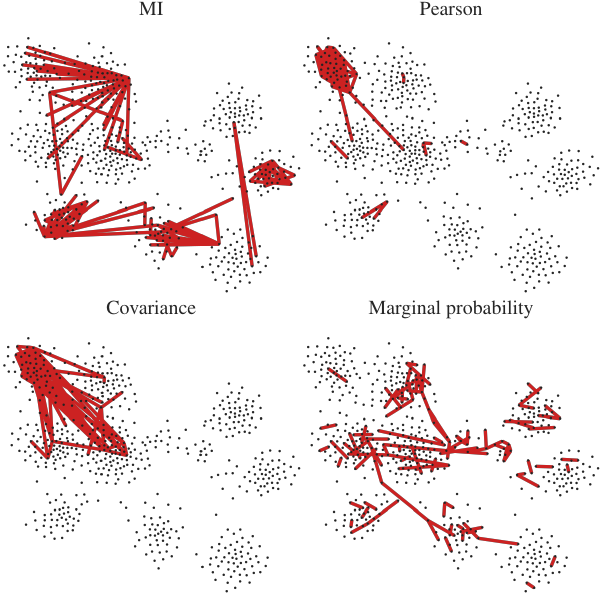}\\
    \includegraphics[width=\columnwidth]{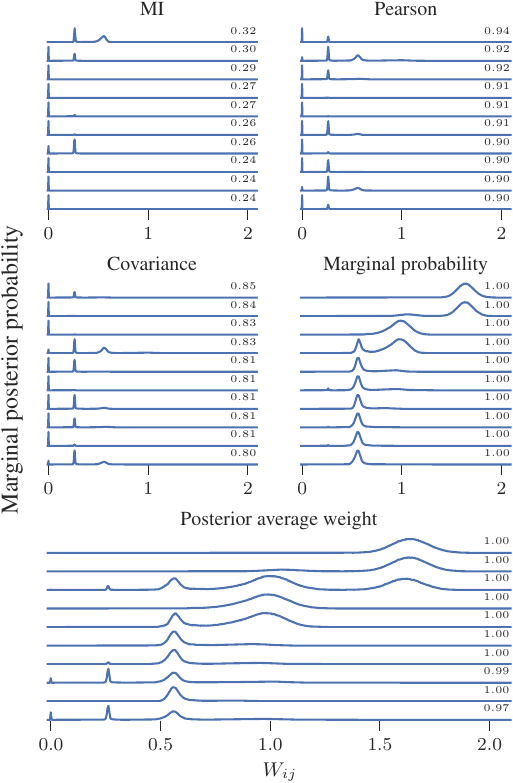}
  \end{tabular}
  \caption{Top: First 100 edge pairs with the largest values of mutual
    information, Pearson correlation, covariance, and marginal probability, for
    the Brazilian congress data. The layout of the nodes is the same as in
    Fig.~\ref{fig:camara}. Bottom: Marginal weight distribution of the 10
    highest ranking node pairs according to the same scores as in the top panel,
    as well as the posterior average weight. The upper right corners show the
    corresponding scores. \label{fig:edge_post}}
\end{figure}

\section{Empirical dynamics}\label{sec:empirical}

In order to investigate the uncertainty information that posterior sampling can
provide for network reconstruction, we first consider the voting dynamics in the
lower house of the Brazilian congress, during the legislative period from 2007
to 2011, involving 623 deputies who voted ``no,'' ``abstain,'' or ``yes'' on 619
voting sessions. We modelled this dynamics according to an equilibrium Ising
model, modified to include the states $\{-1,0,1\}$, corresponding, respectively,
to the aforementioned vote outcomes. The results are shown in
Fig.~\ref{fig:camara}.

The reconstruction uncovers a network ensemble that is divided in 11 groups of
nodes who tend to vote in similar ways. As shown in Fig.~\ref{fig:meta}, the
divisions coincide very well with known party affiliations. The existence of
nonzero couplings between deputies have uncertainties that vary in the entire
$\pi_{ij}\in [0,1]$ range, indicating a very heterogeneous mixture of certain
and uncertain edges. The coupling strengths themselves are distributed around
four typical values, whereas the node biases are centered closely around a
typically small, but positive value, indicating that deputies have only a very
small tendency to vote ``yes'' in the absence of any interaction with their
neighbors. The increased accuracy that the marginal estimate provides is
noticeable when compared to the MAP estimate of Fig.~\ref{fig:camara}c, for
which only 8 groups can be identified, with three groups in the government
coallition being merged together (corresponding to the four groups in the upper
left of Fig.~\ref{fig:camara}b). The tenuous intra-coalition organization is
only visible when the more detailed analysis from posterior sampling is
performed, and implies that the observed dynamics cannot be well captured by a
single network---at least not with the dynamical model used. The similarity
between both estimates is $s(\widehat\W,\W^{*}) = 0.72$, showing that, while
there is a substantial agreement between both estimates, the disagreement is not
negligible (unlike the sufficient data limit in Fig.~\ref{fig:synthetic}), and
indicates how posterior sampling can be important to uncover uncertainties in
the analysis of empirical data.

Our approach allow us to query the individual marginal distributions
$P(W_{ij}|\X)$ for every pair $(i,j)$, giving a substantial amount of
information on the reconstruction, when compared to the MAP point estimate, as
can be seen in Figs.~\ref{fig:camara}e and f.

We move now to another, larger dataset composed of $M=2516$ log-returns of
$N=6369$ stocks in the US market, corresponding to 10 years from 2014 to 2024,
obtained from Yahoo finance~\footnote{Retrieved from the API to
  \url{https://finance.yahoo.com}}. We performed a reconstruction using a
multivariate Gaussian distribution (see Appendix~\ref{app:models}), with $\W$
corresponding to the precision matrix, so that if $W_{ij} = 0$ it means that $i$
and $j$ are conditionally independent. The results are shown in
Fig.~\ref{fig:stock}. Similarly to the previous example, the reconstruction
uncovers a modular network, with edge uncertainties spanning a wide range. As
seen in Fig.~\ref{fig:meta}, the groups found correlate moderately with the
industry sector, although not as clearly as the correlation with party
affiliation in the Brazilian congress example, considered previously. In this
case, the correspondence between the MP and MAP estimates is higher, with a
similarity $s(\hat\W,\W^{*})=0.83$, but the discrepancy is still not negligible,
indicating a somewhat more concentrated posterior distribution (this can also be
seen in Fig.~\ref{fig:stock}b, which shows a larger abundance of edges with
$\pi_{ij}\approx 1$).

\subsection{Comparison between posterior probabilities, weight magnitudes, and pairwise correlations}

We take the opportunity to compare the outcome of our probabilistic
reconstruction with commonly used heuristics for this task, based on pairwise
correlations between the observable behavior of nodes. The biggest disadvantage
of this type of heuristic is the conflation it makes between direct and indirect
neighbors, since if two connected nodes have a high correlation value, the same
is also likely to be true between one of the endpoints involved and any of the
neighbors of the other endpoints. For example, for any three vectors $\bm x$,
$\bm y$, and $\bm z$, the Pearson correlation coefficient must fulfill
\begin{multline}
  \operatorname{corr}(\bm x, \bm z) \ge \operatorname{corr}(\bm x, \bm y) \operatorname{corr}(\bm y, \bm z) \\
  - \sqrt{[1-\operatorname{corr}(\bm x, \bm y)^{2}][1-\operatorname{corr}(\bm y, \bm z)^{2}]}.
\end{multline}
So, e.g. if
$\operatorname{corr}(\bm x, \bm y) = \operatorname{corr}(\bm y, \bm z) = .99$,
then $\operatorname{corr}(\bm x, \bm z) \ge 0.96$, regardless if $\bm x$ and
$\bm z$ correspond to nodes that are truly connected or not. Since the
covariance is related to the Pearson correlation via
$\operatorname{corr}(\bm x, \bm z) = \operatorname{cov}(\bm x, \bm y)/\sqrt{\operatorname{cov}(\bm x, \bm x)\operatorname{cov}(\bm y, \bm y)}$,
the same kind of inherent constraint also affects it. Similarly, mutual
information satisfies
\begin{equation}
\operatorname{MI}(\bm x, \bm z) \ge \operatorname{MI}(\bm x,\bm y) + \operatorname{MI}(\bm y, \bm z) - H(\bm y),
\end{equation}
where $H(\bm y)$ is the entropy of $y$. So, if
$\operatorname{MI}(\bm x,\bm y) = \operatorname{MI}(\bm y, \bm z) = H(\bm y) - \epsilon$,
then $\operatorname{MI}(\bm x, \bm z) \ge H(\bm y) - 2\epsilon$. Therefore, the
idea of simply thresholding these quantities cannot be reconciled with the
distinction between direct and indirect neighbors, at least not in the general
case. This contrasts markedly with the inferential approach considered in this
work, for which such inherent constraints are inexistent.

Nevertheless, we might posit that there are situations where these
reconstruction approaches yield similar results. For example, for a sparse,
homogeneous true network, with all edges having the exact same weight, and all
nodes having the same degree---such that the observed correlation between all
true neighbors is approximately the same---it could be that the small drop in
correlation between first and second neighbors is sufficient to discriminate
between true and false edges.

In order to investigate the quantitative discrepancies outside such an idealized
scenario, in Fig.~\ref{fig:scatter} we show the correspondence between either
the inferred weights or marginal edge probabilities and the three aforementioned
correlation functions, for the two datasets considered so far. In all cases,
although some positive correlations can be detected, they are very weak, meaning
that these correlations are very inefficient predictors of both the presence of
an edge and its weight magnitude. Importantly, the lack of correspondence occurs
even at the extremes: we very often observe edge pairs with close to maximal
correlation, but which nevertheless have a close to zero marginal edge
probability, and conversely, nodes with very high marginal probability or
inferred weight, but which have very low correlation values. This demonstrates
that the inferences that are obtained via our reconstruction approach are
leveraging much more nuanced information in the data than simply whether the
pairwise node correlations are either large or small.

Incidentally, we also investigated the correlation between the inferred weights
and edge probabilities. Naively, one might expect that a large inferred weight
magnitude is synonymous with a large marginal probability, but in reality the
situation is more nuanced. It can be, for example, that a node accepts two other
nodes as equally plausible neighbors with high weight magnitudes, but \emph{not
  simultaneously}, i.e. it is either one node or the other, but not both. In
this case, each of those edges will have a large weight, but a marginal
posterior probability of only 50\%. As can be seen in Fig.~\ref{fig:scatter_wp},
in the case of the Brazilian congress data we do observe a positive correlation
between weight and marginal probability, but it becomes significantly weaker
above $\pi_{ij}=1/2$, meaning that while a sufficiently low weight magnitude
implies low probability, large weights do not necessarily have correspondingly
high probabilities. On the other hand, for the US stocks data the correlation
variance is much stronger, meaning that, while on average a larger weight
implies higher probability, there is an abundance of exceptions, even at the
extremes.

When we compare the reconstructed networks using correlation thresholds with the
inferred ones, as we might expect from the above analysis, we obtain extreme
discrepancies. In Fig.~\ref{fig:corr_acc} we show the Jaccard similarity between
the threshold-based reconstructions and the marginal probabilities for the
Brazilian congress data, which peaks at around $0.16$ for the Pearson
correlation, representing the closest result overall. Even when considering only
the true positive rate---which ignores the inclusion of spurious edges (false
positives) in the reconstruction---the maximum value reaches only similar low
ranges. Importantly, the different correlation functions also disagree
significantly between themselves, as can be seen in Fig.~\ref{fig:edge_post},
which shows the highest scoring node pairs in each case. The same figure also
shows the marginal posterior distribution of weights for the same pairs,
illustrating the lack of agreement between high correlation among nodes and the
weights inferred.

From these comparisons we can conclude that posterior sampling not only provides
valuable uncertainty quantification, but also a completely different, and more
accurate, reconstruction result than comparatively crude, but often employed
heuristics based on thresholding of correlations.

\section{Conclusion}\label{sec:conclusion}

We have described an efficient method to sample from posterior distributions of
networks that allows us to perform uncertainty quantification for the problem of
network reconstruction, as well as to produce consensus estimates from marginal
distributions.

Our method does not rely on specific properties of particular generative models
used for reconstruction, nor on the prior distribution used for their
parameters. We showed how our method can be used together with a sophisticated
regularization scheme that uncovers the most appropriate number of edges and
weight distribution in a manner consistent with the statistical evidence
available in the data.

We have demonstrated on synthetic and empirical examples how posterior sampling
can improve the accuracy of network reconstructions, and uncovers the entire
range of possible reconstructions weighted according to their plausibility as an
account of how the data has been generated.

A comparison with heuristics based on the thresholding of pairwise correlations
revealed the relative advantage of performing an inferential reconstruction,
since besides providing a generative model, uncertainty estimates, and
significantly increased accuracy, it is able of distinguishing between the
probability of existence of an edge and its weight magnitude, which otherwise
would be conflated.

Since our methodology is easily adaptable to other generative models, it remains
to be explored how it can be employed with models more realistic than the
relatively simple ones considered here, and how the underlying Bayesian
framework can be leveraged to perform model selection, to investiagate the
fundamental limits of network reconstruction, and to obtain predictive
statements about the unseen behavior and the outcome of interventions in network
systems, based solely on indirect non-network data.

\bibliography{bib}

\appendix

\section{MDL regularization and joint SBM inference}\label{app:mdl}

Following Ref.~\cite{peixoto_network_2024} we consider a formulation of the edge
weight priors based on a sparse, adaptive quantization of the allowed values,
which amounts to an implementation of the minimum description length (MDL)
principle. More specifically, we first sample an auxiliary unweighted multigraph
$\A$, specifying the placement of nonzero weights, according to the
degree-corrected stochastic block model (DC-SBM)~\cite{karrer_stochastic_2011},
here in its microcanonical formulation~\cite{peixoto_nonparametric_2017}, with a
likelihood
\begin{equation}
  P(\A |\bb,\bm k,\bm e) = \frac{\prod_{r<s}e_{rs}!\prod_{r}e_{rr}!!\prod_{i}k_{i}!}{\prod_{i<j}A_{ij}\prod_{i}A_{ii}!!\prod_{r}e_{r}!},
\end{equation}
where $\bb=\{b_{i}\}$ is the node partition, with $b_{i}\in\{1,\dots,B\}$ being
the group membership of node $i$, $\bm k =\{k_{i}\}$ is the degree sequence,
with $k_{i}$ being the degree of node $i$, and $\bm e = \{e_{rs}\}$ is the group
affinity matrix, with $e_{rs}$ being the number of edges between groups $r$ and
$s$, or twice that if $r=s$. Based on the multigraph $\bm A$, a simple graph
$\bm G$ is obtained by ``erasing'' the edge
multiplicities~\cite{peixoto_latent_2020},
\begin{equation}
  P(\bm G | \A) = \prod_{i<j}{(1-\delta_{A_{ij},0})}^{G_{ij}}{\delta_{A_{ij},0}}^{1-G_{ij}}.
\end{equation}
Conditioned on $\bm G$, we sample the nonzero weights from a finite set of $K$
values $\bm z = \{z_{1},\dots,z_{K}\}$, conditioned on their exact counts
$\bm m = \{m_{k}\}$, where $m_{k}=\sum_{i<j}\delta_{W_{ij}, z_{k}}$, and
otherwise uniformly, according to
\begin{multline}
  P(\W|\bm z, \bm m, \bm G)
  = \frac{\prod_{k}m_{k}!}{E!} \\ \times \prod_{k}\delta_{m_{k},\sum_{i<j}\delta_{W_{ij},z_{k}}} \times \prod_{i<j}\delta_{W_{ij},0}^{1-G_{ij}},
\end{multline}
with the nonzero counts themselves sampled uniformly according to
\begin{equation}
  P(\bm m | K, \bm A)=\frac{\delta_{\sum_{k}m_{k},E(\A)}\prod_{k}\mathds{1}_{m_{k}>0}}{{E(\A)-1 \choose K-1}},
\end{equation}
where $E(\A)$ is the number of nonzero entries in $\bm A$.
In Ref.~\cite{peixoto_network_2024} the weight categories were sampled according
to a discrete Laplace distribution. Instead, here we propose a slight variation,
where only the extreme values $z_1$ and $z_K$ are sampled
jointly as
\begin{multline}
  P(z_1, z_K|\lambda,\Delta) =
  \mathds{1}_{z_1 \le z_K}(2-\delta_{z_1, z_K})  \\\times P(z_1|\lambda,\Delta)P(z_K|\lambda,\Delta),
\end{multline}
where
\begin{multline}\label{eq:z}
  P(z|\lambda,\Delta)\\
  \quad=\begin{cases}
      \ee^{-\lambda|z|}(\ee^{\lambda\Delta} - 1)/2, &\text{ if }\; \begin{aligned}
                                                                          z&=\Delta\left\lceil z/\Delta\right\rfloor, \text{ and} \\
                                                                          z&\neq 0
                                                                        \end{aligned}\\
      0, &\text{ otherwise, }
   \end{cases}
 \end{multline}
 is a quantized zero-excluded Laplace distribution, with decay and quantization
 parameters, $\lambda$ and $\Delta$, respectively, each sampled uniformly from
 the set of all strictly positive real numbers representable by $q$ bits, i.e.
 $P(\Delta|q) = P(\lambda|q) = 2^{-q}$, for which we pragmatically choose
 $q=64$. Conditioned on these extreme values, we sample the remaining $K-2$
distinct values uniformly as
\begin{multline}
  P(z_{2},\dots,z_{K-1} | \Delta, K, z_1, z_K) = \\
  \frac{\prod_{k=2}^{K-1}\delta_{z_{k}, \Delta\floor{z_{k}/\Delta}}\times\prod_{k=1}^{K-1}\mathds{1}_{z_{k} < z_{k+1}}}{{{(z_K - z_1)/\Delta - 1 - \mathds{1}_{z_1z_K < 0}} \choose {K-2}} }.
\end{multline}
Lastly, the number $K$ of discrete weight values is sampled uniformly inside the
allowed range according to
\begin{equation}
  P(K|\Delta, z_1, z_K) =
  \frac{\mathds{1}_{1 \le K \le  (z_K - z_1)/\Delta + 1 - \mathds{1}_{z_1z_K < 0}}}
  {(z_K - z_1)/\Delta + 1 - \mathds{1}_{z_1z_K < 0}}.
\end{equation}
Putting it all together we have
\begin{widetext}
\begin{align}
  P(\W|\A,\lambda,\Delta)
  &= \sum_{\bm G}\left[\sum_{\bm z, K}P(\W|\bm z, \bm G)P(z_{2},\dots,z_{K-1} | \Delta, K, z_1, z_{K})P(K|z_{1},z_{K},\Delta) P(z_1, z_K|\lambda,\Delta)\right]^{1-\delta_{E(\bm G),0}} P(\G|\bm A) \nonumber \\
         &=\left[\frac{\left(\prod_{k}m_{k}!\right) \ee^{-\lambda (|z_1| + |z_K|)}(\ee^{\lambda\Delta} - 1)^{2}(2-\delta_{z_1, z_K})}{4\times E!{E-1 \choose K-1}{{(z_K - z_1)/\Delta - 1 - \mathds{1}_{z_1z_K < 0}} \choose {K-2}} \left[(z_K - z_1)/\Delta + 1 - \mathds{1}_{z_1z_K < 0}\right]} \right]^{1-\delta_{E(\A),0}}\prod_{i<j}\delta_{W_{ij},0}^{\delta_{A_{ij}, 0}},\label{eq:W_prior}
\end{align}
\end{widetext}
where the remaining quantities $\bm m$, $\bm z$, $K$, and $E$ in
Eq.~\ref{eq:W_prior} should be interpreted as being functions of $\bm W$.

With this prior at hand, we can formulate the problem of
reconstruction according to the joint posterior
\begin{multline*}
  P(\W, \A, \bb,\lambda,\Delta | \X) = \\
  \frac{P(\X|\W)P(\W|\A,\lambda,\Delta)P(\A|\bb)P(\bb)P(\lambda)P(\Delta)}{P(\X)},
\end{multline*}
where the marginal distribution
$P(\A|\bb) = \sum_{\bm k, \bm e, E}P(\A |\bb,\bm k,\bm e)P(\bm k|\bm e)P(\bm e | E)P(E)$
is computed using the priors described in
Ref.~\cite{peixoto_nonparametric_2017}, in particular those corresponding to the
hierarchical (or nested) SBM~\cite{peixoto_hierarchical_2014}. The prior for the
total number of mutiedges, $P(E)=[\mu/(\mu + 1)]^{E}/(\mu + 1)$, is a geometric
distribution with mean $\mu = {N \choose 2}$, and comparable standard deviation
$\sigma_{E}=\sqrt{\mu(\mu + 1)} \approx {N \choose 2}$, for $N\gg 1$.

The proposals for the partitions $\bb$ are done according to the merge-split
algorithm described in Ref.~\cite{peixoto_merge-split_2020}. Although it is
straightforward to introduce move proposals for both $\lambda$ and $\Delta$, we
found that the results are often indistinguishable from simply choosing
$\lambda=1$ and $\Delta = 10^{-8}$, since these are not very sensitive
hyperparameters.

For generative models which have additional node parameters, e.g. local
fields of the Ising model (see Appendix~\ref{app:models}), almost identical
priors can be used for them, with the only exception being that zero values are
allowed. See Ref.~\cite{peixoto_network_2024} for details.

\section{Edge weight proposals via bisection and linear interpolation (BLI)}\label{app:sample_x}

In the main text we focused on selecting which node pairs to update, but gave no
details about how the edges should be updated, i.e. what should be the move
proposal $Q(W_{ij}'|\W, i, j)$ after we have selected the node pair $(i,j)$. A
standard approach in this case would be to choose for this a normal distribution
centered on the previous value, and with some user-defined variance. However,
this has as a drawback that the variance needs to be carefully chosen, which in
general requires a substantial degree of experimentation and fine-tuning. Here
we describe an alternative bisection and linear interpolation (BLI) approach,
that is self-adaptive and does not require fine-tuning. We start with a triplet
$(W_{a}, W_{b}, W_{c})$, with $W_{a} < W_{b} < W_{c}$, that ``brackets'' a
maximum in the conditional posterior $f(W) = P(W_{ij}=W|\W\setminus W_{ij},\X)$,
i.e.
\begin{equation}
  f(W_{a}) < f(W_{b}), \qquad f(W_{b}) > f(W_{c}).
\end{equation}
If this condition is fulfilled, then there is at least one local maximum in the
interval $[W_{a},W_{c}]$. Such a triplet can be found by considering an initial
$(W_{a}^{\text{init}}, y, W_{c}^{\text{init}})$, with $W_{a}^{\text{init}}$ and
$W_{b}^{\text{init}}$ being initial guesses that bound the typical range of
weight values, and $y$ is sampled uniformly at random in the interval enclosed
by these values. If this initial choice does not bracket a maximum, the boundary
$W$ with the largest $f(W)$ is multiplied by a factor 2. This procedure is
repeated until a bracketing interval is found, and the difference between\
$\log f(W_{b}) - \log \max(f(W_{a}),f(W_{c}))$ is sufficiently large, e.g.\ more
than 200 or so, such that values outside this range can be neglected as having a
vanishingly small probability. Having obtained this bracketing interval, we
proceed with a random bisection search:
\begin{enumerate}
  \item We sample $y$ uniformly at random between either
        $[W_{a},W_{b}]$ or $[W_{b},W_{c}]$, depending on which interval is larger.
  \item The new bracketing interval is updated to include $y$ as its midpoint
        and the old midpoint $W_{b}$ as one of the boundaries if
        $f(y) > f(W_b)$, otherwise the midpoint is preserved and the
        corresponding boundary is updated to $y$.
  \item If $\log f(W_{b}) - \log \max(f(W_{a}),f(W_{c})) < \epsilon$, the search
        stops. Otherwise we go back to step 1.
\end{enumerate}
The above algorithm will converge to a \emph{local} maximum of $f(W)$ after
$O(\log(1/\epsilon))$ iterations on average. The fact we select the midpoint
uniformly at random---instead of deterministically like in the golden section
search method~\cite{press_numerical_2007}---means we can in principle obtain any local maximum contained
in the initial interval.

\begin{figure}
  \begin{tabular}{cc}
    \begin{overpic}[width=.5\columnwidth]{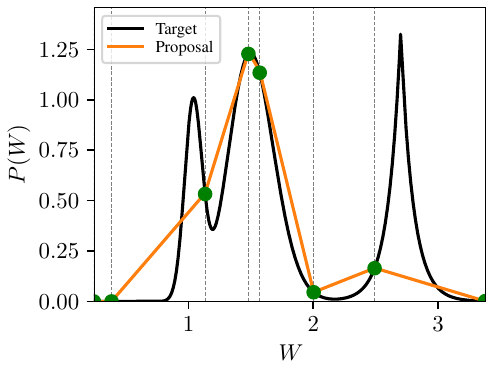}\put(0,73){(a)}\end{overpic}&
    \begin{overpic}[width=.5\columnwidth]{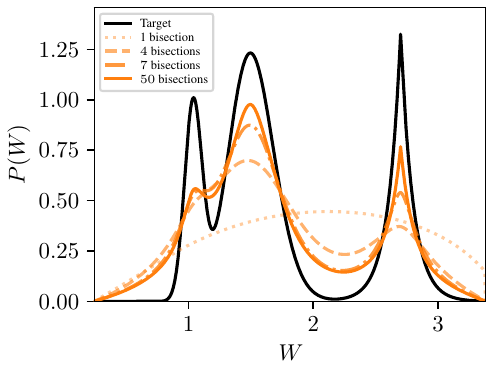}\put(0,73){(b)}\end{overpic}\\
    \multicolumn{2}{c}{\begin{overpic}[width=\columnwidth]{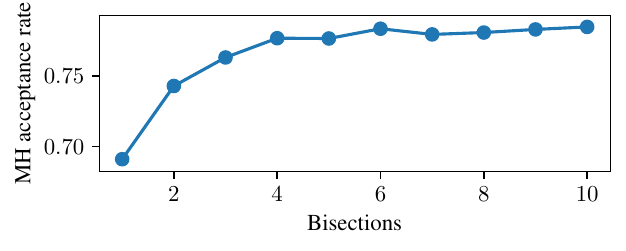}\put(0, 42){(c)}\end{overpic}}
  \end{tabular}
  \caption{(a) Example target distribution and the proposal generated via the
    algorithm described in the main text. The circle markers and the vertical
    lines mark the random bisection points. (b) Average proposal distribution
    for increasing number of bisection steps, as shown in the legend. (c)
    Metropolis-Hastings (MH) acceptance rate as a function of the number of
    bisections.\label{fig:wproposal}}
\end{figure}

Our objective is to produce a sample proposal from $f(W)$, not to optimize it.
So we construct a distribution formed by a linear interpolation between all the
points considered during the random bisection algorithm above, which by
necessity involves the neighborhood of at least one local maximum, and therefore
probes regions of relative high probability from the target distribution. This
interpolation requires a number of points $n=O(\log(1/\epsilon))$, and a single
sample from it can be generated in time $O(n)$, by first computing the relative
probability mass for each linear segment, then sampling a linear segment
according to these probabilities (e.g.\ with the alias
method~\cite{walker_new_1974,vose_linear_1991}, requiring time $O(n)$), and
finally sampling the final value inside the interval in time $O(1)$ by an
inverse transform. An example run of this scheme is shown in
Fig.~\ref{fig:wproposal} for a multimodal target distribution. As can be seen in
Fig.~\ref{fig:wproposal}b, which shows an average of many such proposals, the
proposals tend to concentrate around the modes of the target distribution, and,
in this example, more than 4 bisections does not bring noticeable
improvements---therefore only very few likelihood evaluations are needed. In
Fig.~\ref{fig:wproposal}c is also shown the average Metropolis-Hastings (MH)
acceptance rate as a function of the number of bisections, demonstrating the
same saturation at around 4 bisections for this particular example.

For the specific generative models consided in the main text and in
Appendix~\ref{app:models}, their corresponding conditional likelihood $f(W)$ is
convex, which means that a deterministic bisection could be used instead.
However, in the interest of generality, our algorithm does not rely on the
convexity of the conditional likelihood, nor on other usually desirable
properties such as it being differentiable or even continuous.

We note also that when computing the MH acceptance probability,
it is not necessary to include the probability of choosing the bisection points
themselves, nor the marginal probability averaged over all of them. We notice
this by considering the detailed balance condition
\[
  f(W')T(W'|\bm\gamma)P(\bm\gamma) = f(W)T(W|\bm\gamma)P(\bm \gamma),
\]
with $\bm \gamma$ being the random bisection points chosen with the above
algorithm. If this condition is fulfilled, then the marginal detailed balance is
also trivially fulfilled, i.e. $f(W')T(W') = f(W)T(W)$, with
$T(W)=\int T(W|\bm\gamma)P(\bm \gamma)\dd\bm\gamma$, and the MH acceptance is
computed as
\[
  \begin{aligned}
    a &= \min\left(1, \frac{f(W')P(W|\bm\gamma)P(\bm\gamma)}{f(W)P(W|\bm\gamma)P(\bm\gamma)} \right)\\
      &= \min\left(1, \frac{f(W')P(W|\bm\gamma)}{f(W)P(W|\bm\gamma)} \right).\\
  \end{aligned}
\]
which is independent of $P(\bm\gamma)$, and depends only on the probability
$P(W|\bm\gamma)$ of sampling the final value according to the bisection points
$\bm\gamma$, which is easily computed from the linear interpolation.

\subsection{Discrete values}

When dealing with the discretized values for $\W$ considered in
Appendix~\ref{app:mdl}, special considerations are needed. Although we can
easily adapt the above BLI sampling to values which are multiples of the
quantization parameter $\Delta$, this may not yield proposals which are
accepted, since most of the time the proposal will yield a new value of $z_{k}$,
increasing the number $K$ of discrete categories, which, per design, exerts a
penalty to the likelihood. Becase of this, we consider the following move types:
\begin{enumerate}
  \item New categories: BLI moves constrained to values which are multiples of $\Delta$.
  \item Old categories: BLI moves constrained to the existing categories, $\bm z$.
  \item Collective category moves: BLI moves of a single category $z_{k}$ with
        $k\in \{1,\dots,K\}$, to a new value which is a multiple of $\Delta$,
        distinct from the other categories.
\end{enumerate}
Move types 1 and 2 are mutually required to fulfill detailed balance, since, if
the current category has more than one count, the move to a new category can
only be reversed by a move to a previously existing category, and vice versa for
the vanishing of an existing category with a single count. Move type 3 will
simultaneously involve all the edges that belong to the same category, and thus
can be seen as a non-local move that can speed up the MCMC convergence, and is
an inherent advantage offered by our discretized approach.

Furthermore, we also employ the merge-split of
Ref.~\cite{peixoto_merge-split_2020} for the distribution of the weight
categories on the edges, since this can remove likelihood barriers that exist
when moving one edge at a time. The only modification we use for that algorithm
is that when weight categories are split and merged, the respective category
values $z_{k}$, both for old and new categories, are sampled according to the
BLI algorithm described previously.

\section{Updating multiple entries simultaneously: edge replacements and swaps}\label{app:swap}

The move proposals considered in the main text all involve the update of a
single entry of the matrix $\W$ at a time. In the presence of non-convex
regularization schemes that penalize the excessive abundance of edges, we can
encounter scenarios were the respective removal and addition of edges in two
different entries of $\W$ would be individually rejected, but if these are
performed at the same time their combined move would be accepted. In this way,
the regularization can introduce ``barriers'' in the posterior landscape that
slow down the mixing of the Markov chain. In order to avoid this, here we
consider also updates that involve two entries simultaneously. The first type of
move is an edge replacement, performed as follows:
\begin{enumerate}
  \item A node $i$ is sampled uniformly at random.
  \item A neighbor of $j$ is sampled uniformly at random with probability
    \[
    P_{e}(j|i) =
    \begin{cases} A_{ij}/\sum_uA_{iu}, & \text{ if } \sum_uA_{iu} > 0,\\
      1/N, & \text{ otherwise, }
    \end{cases}
    \]where we account also for nodes with degree zero.
  \item A node $v$ is sampled with probability $P_f(v|i)$.
  \item If $|\{i,j,v\}| < 3$, i.e. at least one of the nodes is repeated, the
        proposal is skipped.
  \item Otherwise, the values of the entries $W_{ij}$ and $W_{iv}$ are swapped.
\end{enumerate}
In the above, the new potential neighbor is sampled in step 3 with probability
\begin{equation}
  P_f(j|i) = p P_{\hat{\mathcal{E}}}(j|i) + (1-p)P_{\Lambda}(j|i),
\end{equation}
where $p$ is the probability of sampling according to the typical edge set
estimate, i.e.
\begin{equation}
P_{\hat{\mathcal{E}}}(j|i) =
    \begin{cases} G_{ij}/\sum_uG_{iu}, & \text{ if } \sum_uG_{iu} > 0,\\
      1/N, & \text{ otherwise, }
    \end{cases}
\end{equation}
where $\bm G$ is the adjacency matrix corresponding to the edges in
$\hat{\mathcal{E}}$, or otherwise they are sampled according to the nodes
reachable from $i$, i.e. with probability
\begin{equation}
  P_\Lambda(j|i) =
  \begin{cases}
    q\frac{\mathds{1}_{j \in \Lambda(i, d)}}{|\Lambda(i, d)|} + \frac{1-q}{N}, \text{ if } |\Lambda(i, d)| > 0,\\
    \frac{1}{N}, \text{ otherwise,}
  \end{cases}
\end{equation}
and where $\Lambda(i,d)$ is the set of nodes reachable from $i$ at a
distance at most $d$, and $1-q$ is the probability of choosing $v$ uniformly
at random.

Note that the above proposal will either move an edge from $(i,j)$ to $(i, v)$,
if $W_{iv}=0$, or simply swap their weights otherwise. However, if a move is
performed, it will change the number of neighbors of $v$ and $j$. Because of
this, we can also consider a swap proposal that can preserve the degrees of all
nodes involved, namely, we select four nodes $\{i,j,u,v\}$ according to
\begin{equation}
  P(i,j,u,v) = P(i)P_e(j|i)P_f(u|j)P_e(v|u),
\end{equation}
and if $|\{i,j,u,v\}| < 4$, i.e. at least one of the nodes is repeated, we skip
the proposal, otherwise we swap $W_{ij}$ with $W_{iv}$, and $W_{uv}$ with
$W_{uj}$. Note that this will preserve the node degrees only if $W_{ij}$ and
$W_{uv}$ are both nonzero, and $W_{iv}$ and $W_{uj}$ are both zero.

We do not analyze the effect of these move proposals in detail, but they are
included in our reference implementation, and we have observed a positive effect
in the mixing time of empirical networks.

\section{Generative models}\label{app:models}

In our examples we use three generative models: the equilibrium Ising
model~\cite{nguyen_inverse_2017}, the kinetic Ising model, and a multivariate
Gaussian.

The kinetic Ising model is a Markov chain on $N$ binary variables
$\bm{x} \in \{-1,1\}^{N}$ with transition probabilities given by
\begin{equation}\label{eq:ising_kinetic}
  P(\bm x(t+1) |\bm x(t), \W, \bm \theta) =
  \prod_{i}\frac{\ee^{x_{i}(t+1)(\sum_{j}W_{ij}x_{j}(t) + \theta_i)}}{2\cosh(\sum_{j}W_{ij}x_{j}(t) + \theta_i)},
\end{equation}
with $\theta_{i}$ being a local field on node $i$.

The equilibrium Ising model is the $t\to\infty$ limiting distribution of the
above dynamics, with a likelihood given by
\begin{equation}\label{eq:ising}
  P(\bm x | \W, \bm \theta) = \frac{\ee^{\sum_{i<j}W_{ij}x_{i}x_{j} + \sum_{i}\theta_{i}x_{i}}}{Z(\W, \bm \theta)},
\end{equation}
with $Z(\W,\bm\theta)=\sum_{\bm x}\ee^{\sum_{i<j}W_{ij}x_{i}x_{j} + \sum_{i}\theta_{i}x_{i}}$
being a normalization constant. Since this normalization cannot be computed
analytically in closed form, we make use of the pseudolikelihood
approximation~\cite{besag_spatial_1974},
\begin{align}
  P(\bm x | \W, \bm \theta) &= \prod_{i}P(x_{i}|\bm x\setminus {x_{i}}, \W, \bm \theta)\\
  &= \prod_{i}\frac{\ee^{x_{i}(\sum_{j}W_{ij}x_{j} + \theta_i)}}{2\cosh(\sum_{j}W_{ij}x_{j} + \theta_i)},\label{eq:ising_pseudo}
\end{align}
---which essentially approximates Eq.~\ref{eq:ising} as the probability of a
transition of the global state of the kinetic Ising model onto itself---since it
gives asymptotically correct results and has excellent performance in
practice~\cite{mozeika_consistent_2014,nguyen_inverse_2017}.

In the case of the zero-valued Ising model with $\bm{x}\in \{-1,0,1\}^{N}$, the
normalization of Eqs.~\ref{eq:ising_pseudo} and~\ref{eq:ising_kinetic} change
from $2\cosh(\cdot)$ to $1+2\cosh(\cdot)$.

Finally, the (zero-mean) multivariate Gaussian is a distribution on
$\bm{x} \in \mathbb{R}^{N}$ given by
\begin{equation}
  P(\bm x | \bm W) = \frac{\ee^{-\frac{1}{2} {\bm x}^{\top}\W \bm x}}{\sqrt{(2\pi)^N |\bm W^{-1}|}},
\end{equation}
where $\bm W$ is the precision (or inverse covariance) matrix. Unlike the Ising
model, this likelihood is analytical---nevertheless, the evaluation of the
determinant is computationally expensive, and therefore we make use of the same
pseudolikelihood approximation~\cite{khare_convex_2015},
\begin{equation}
  P(\bm x | \bm W, \bm \theta) = \prod_{i}\frac{\ee^{-(x_i + \theta_i^2\sum_{j\neq i}W_{ij}x_{j})^{2}/2\theta_i^2}}{\sqrt{(2\pi)}\theta_i},
\end{equation}
where we parameterize the diagonal entries as $\theta_i=1/\sqrt{W_{ii}}$.

\end{document}